%% file: main.tex
\documentclass[10pt,twocolumn,letterpaper]{article}

\usepackage{iccv}              

\input{preamble}

\definecolor{iccvblue}{rgb}{0.21,0.49,0.74}
\usepackage[pagebackref,breaklinks,colorlinks,allcolors=iccvblue]{hyperref}
\usepackage[accsupp]{axessibility}  
\usepackage{appendix}
\usepackage{colortbl}
\usepackage{xcolor}
\usepackage{graphicx}
\usepackage{array}
\usepackage{float}
\usepackage{amsmath}
\usepackage{amssymb}
\usepackage{mathrsfs}
\usepackage{algorithm}
\usepackage{algpseudocode}
\usepackage{cancel} 
\usepackage{multirow}
\usepackage{adjustbox}
\definecolor{lightblue}{RGB}{220,230,240}

\newcommand{\ModelName}{EEdit}
\title{EEdit\adjustbox{valign=c}{\includegraphics[height=1.4em]{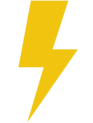}}: Rethinking the Spatial and Temporal Redundancy for \\
Efficient Image Editing}

\author{
Zexuan Yan\textsuperscript{1,*} \quad
Yue Ma\textsuperscript{2,*} \quad
Chang Zou\textsuperscript{1} \quad
Wenteng Chen\textsuperscript{1} \quad
Qifeng Chen\textsuperscript{2} \quad
Linfeng Zhang\textsuperscript{1,\dag} \\
\textsuperscript{1}Shanghai Jiao Tong University \quad 
\textsuperscript{2}Hong Kong University of Science and Technology \\
\textbf{Project: \href{https://eff-edit.github.io/}{\texttt{\textcolor{cyan}{https://eff-edit.github.io/}}}}
}
\begin{document}

\twocolumn[{
\begin{center}
\maketitle
\vspace{-2em}
    \captionsetup{type=figure}
    \includegraphics[width=1.0\textwidth]{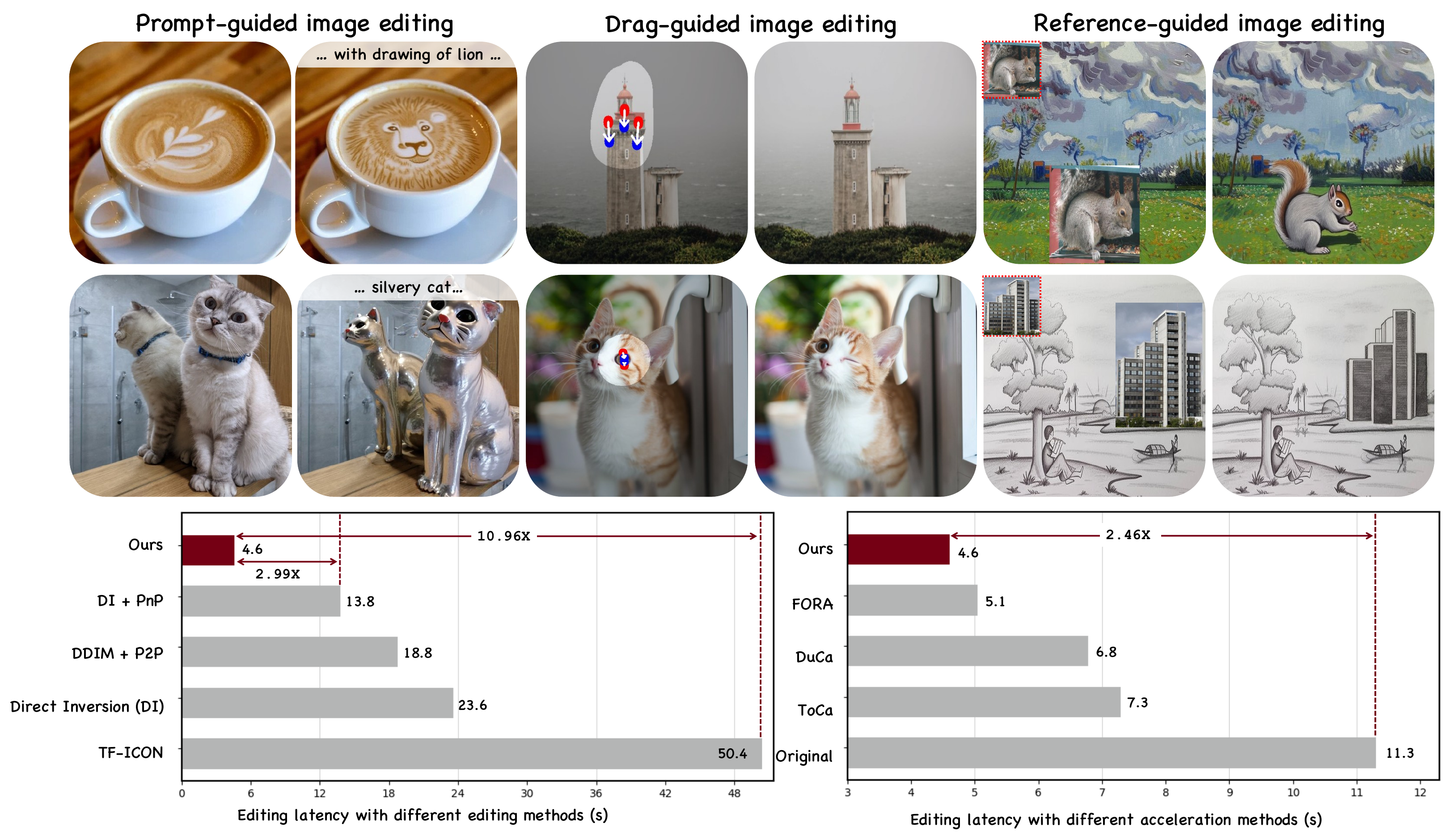}
    \vspace{-2em}
    \caption{\textbf{Gallery of various editing tasks and efficiency comparisons.} We propose the EEdit, a novel inversion-based framework for efficient image editing. Compare with previous methods, we achieve the faster and more efficient image editing. }
\end{center}
}]
\input{sec/0_abstract}    
\input{sec/1_intro}
\input{sec/2_related}
\input{sec/3_preliminaries}
\input{sec/4_approach}
\input{sec/5_exp}
\input{sec/6_conclusion}
{
    \small
    \bibliographystyle{ieeenat_fullname}
    \bibliography{main}
}
\newpage
\input{sec/appendix}

\end{document}

%% file: preamble.tex
\pdfoutput=1


%% file: sec/0_abstract.tex
\noindent
\renewcommand\thefootnote{\fnsymbol{footnote}}
\footnotetext[1]{Contributed equally}
\footnotetext[2]{Corresponding author}
\vspace{-1mm}
\begin{abstract}
\vspace{-1em}
  Inversion-based image editing is rapidly gaining momentum while suffering from significant computation overhead, hindering its application in real-time interactive scenarios. 
  In this paper, we rethink that the redundancy in inversion-based image editing exists in both the spatial and temporal dimensions, such as the unnecessary computation in unedited regions and the redundancy in the inversion progress. 
  To tackle these challenges, we propose an \textbf{E}fficient \textbf{Edit}ing framework, named \textbf{EEdit}, to achieve efficient image editing. Specifically,
  we introduce three techniques to solve them one by one. \textbf{For spatial redundancy}, spatial locality caching is introduced to compute the edited region and its neighboring regions while skipping the unedited regions, and token indexing preprocessing is designed to further accelerate the caching. \textbf{For temporal redundancy}, inversion step skipping is proposed to reuse the latent for efficient editing.
  Our experiments demonstrate an average of \textbf{\textcolor{black}{2.46}}$\times$ acceleration without performance drop in a wide range of editing tasks including prompt-guided image editing, dragging and image composition. Our codes are available at 
  \href{https://github.com/yuriYanZeXuan/EEdit}{\texttt{\textcolor{cyan}{https://github.com/yuriYanZeXuan/EEdit}}}.

\end{abstract}

%% file: sec/1_intro.tex
\begin{figure*}[ht]
    \centering
    \includegraphics[width=1.\textwidth]{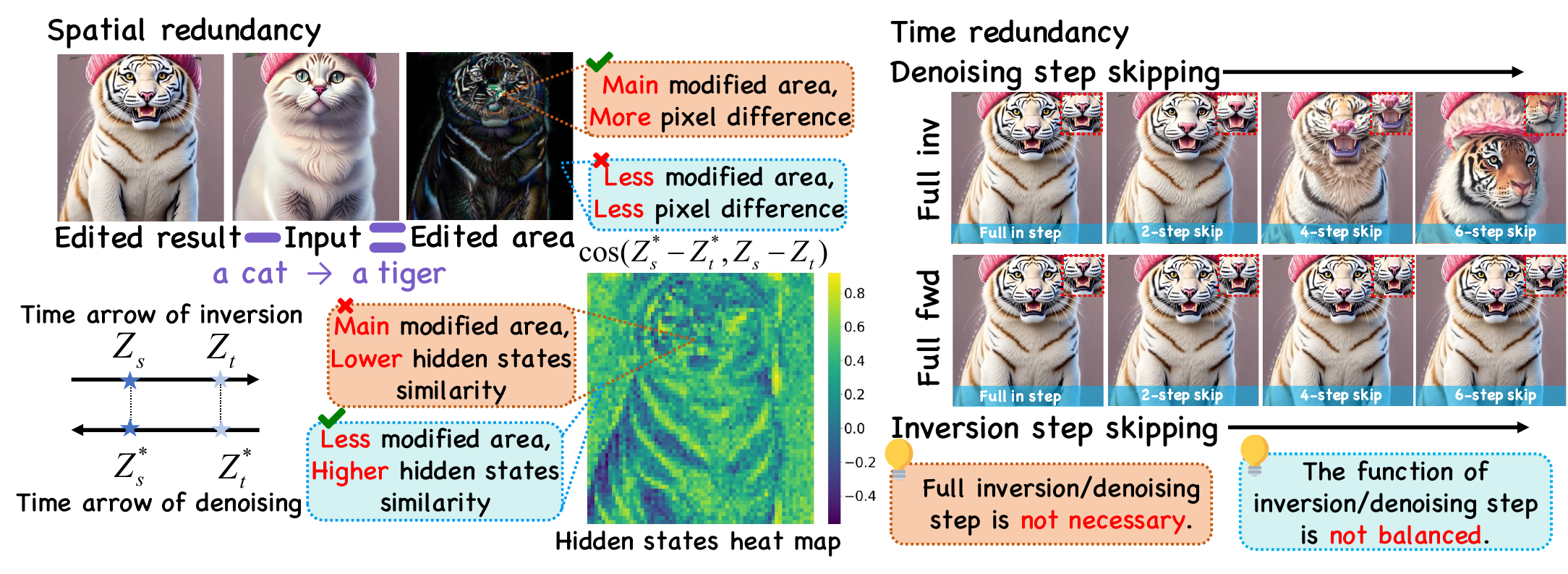}
    \caption{\textbf{The illustration of motivation}. We observe that there are significant computational redundancy during the editing process, which can be classified into spatial redundancy~(a) and temporal redundancy~(b).   For spatial redundancy~(a), due to a high background similarity between the source image and the edited results, calculating the entire image in every inversion/denoising step is unnecessary. For temporal redundancy~(b), we discover that skipping some steps in inversion/denoising does not degrade the editing performance.
    }
    \label{fig:qualitative-intro}
\end{figure*}
\section{Introduction}
\setlength{\abovecaptionskip}{1pt}
\setlength{\belowcaptionskip}{1pt}
With groundbreaking advances in diffusion models~\cite{dit, sd3, Dalva2024FluxSpace, Xie2024SANA}, they have emerged as the state-of-the-art approach for both image generation and editing tasks~\cite{masactrl, pnp, Xu2023InversionFreeEditing, ma2023magicstick, p2p}. Current diffusion-based editing frameworks typically adopt a two-stage pipeline: (1) an \textit{inversion process}~\cite{nulltext, Pan2023AcceleratedDiffusionInversion, ma2024followyouremoji, zhu2024multibooth, chen2024follow, rfinversion, Wang2024TamingRectifiedFlow} that maps the input image to its corresponding noisy latent, followed by (2) a \textit{denoising process}~\cite{Lin2024CommonDiffusionFlaws, Song2022DDIM, tf-icon, wang2024cove, zhu2024instantswap, rfinversion} that progressively denoises and modifies the latent code to produce the edited image.
While achieving impressive editing quality, this two-stage paradigm suffers from significant computational overhead, particularly hindering its usage from real-time interactive applications on resource-constrained edge devices.
To address this problem, this paper begins by identifying two types of computational redundancy in editing and then solves them one by one.

\noindent \textbf{(I) Spatial Redundancy}: Although only a small region of the image is expected to be edited, the current editing pipeline has to compute all the pixels of the given image. Retaining computation in these unedited regions incurs additional computational overhead, while yielding minimal benefits for our editing objectives. In Figure.~\ref{fig:qualitative-intro}, we visualize and compare the differences between pixels before and after image editing, as well as the heat map of cosine similarity of the latent space state differences at fixed time intervals. Both patterns consistently exhibit spatial redundancy differences in the editing task: the edited regions show greater pixel differences and lower similarity in the latent space, whereas the unedited regions exhibit negligible pixel differences and higher similarity in the latent space. Specifically, the unedited region exhibits significantly greater redundancy compared to the edited region.

\noindent \textbf{(II) Temporal Redundancy}: Compared with generating a new image, image editing poses additional computation cost for inversion, which is employed to map the to-be-edited image to the latent space of noise to some extent. In the traditional editing pipeline, the inversion process takes the same computation cost as denoising and thus doubles the computational overhead.

In Figure.~\ref{fig:qualitative-intro}, we separately control the full inversion and denoising processes by introducing interval-skipping time steps during the inversion/denoising process. It allows us to compare how reducing the computational load of one process under the same overall computational cost affects the editing performance. Notably, reducing the denoising steps first led to the loss of fine details, such as texture information around the mouth and nose, and soon resulted in the degradation of overall structural integrity. Surprisingly, skipping time steps in the inversion process had little to no perceptible effect, indicating a highly unbalanced distribution of temporal redundancy between inversion and denoising. Specifically, we discover that inversion exhibits significantly greater redundancy compared to denoising.

Based on this observation, we propose  \textbf{S}patial \textbf{Lo}cality \textbf{C}aching (\textbf{SLoC}), which aims to skip the computation of unedited regions by reusing their features computed in the previous timesteps. Concretely, during both the denoising and inversion process, SLoC first computes the tokens corresponding to all the regions and stores them in a cache. Then, in the following step, SLoC still performs full computation on the edited region and its neighboring regions, while skipping the computation of other regions by reusing their previously cached features. Such a mixed-computation manner in SLoC enables diffusion models to pay more effort in important regions guided by the editing prior, maintaining good quality in the edited region while trading the computation in unedited region for efficiency.
Furthermore, to further reduce the computation from SLoC, 
we analyze that caching initialization and update strategy for
the score map can be finished as an offline operation. Based on this observation, we propose the token index preprocessing, 
achieving over a \textbf{15\%} improvement in inference speed. Importantly, this optimization is mathematically equivalent, ensuring that SLoC itself undergoes an additional lossless acceleration.

Additionally, we observe that there is temporal redundancy in both inversion/denoising processing. Motivated by DDIM~\cite{Song2022DDIM},  we propose the inversion 
step skipping strategy. The primary insight is that \textit
{skipping certain inversion steps does not produce noticeable artifacts and can markedly improve processing speed.} Surprisingly, our experiments reveal that the number of timesteps for inversion can be safely reduced to 33.3\% of that for diffusion, with almost no noticeable performance degradation.

We validated the effectiveness of proposed modules in extensive experiments on various editing tasks, including prompt-guided~\cite{hgan,Dalva2024FluxSpace,pnp}, reference-guided~\cite{tf-icon, dccf, Wang2024PrimeComposer}, and drag-guided editing~\cite{sdedrag,regiondrag,zhao2024fastdrag,shi2023dragdiffusion}. While achieving state-of-the-art performance in background consistency, EEdit also delivers up to \textbf{10.96$\times$} latency acceleration compared to other editing methods. Even when compared to existing cache-based acceleration techniques, it demonstrates a superior acceleration ratio. Furthermore, EEdit achieves a \textbf{2.46$\times$} speedup with minimal performance differences, maintaining near-lossless quality across various evaluation metrics when compared to the original editing pipeline without cache acceleration.

In summary, our contributions include:
\begin{itemize}
    \item We emphasize the spatial and temporal redundancy in inversion-based editing tasks and propose \ModelName, a novel editing framework to modify images efficiently.
    \item \textbf{Spatial Locality Caching~(SLoC):} To reduce the spatial redundancy in editing,  spatial locality caching is proposed to skip most of the computation in unedited regions. Then we design the \textbf{T}oken \textbf{I}ndex \textbf{P}reprocessing~\textbf{(TIP)} to further optimize the speed of caching. 
    \item \textbf{Inversion Step Skipping~(ISS):} To reduce the temporal redundancy, we 
    propose to assign more computation to denoising and fewer computation to inversion, which firstly demonstrates and leverages the unequal importance of the two processes to accelerate editing.
    \item \textbf{Extensive Adaption and Experiments:} We evaluate our methods across various editing tasks, including prompt-guided editing, drag-guided editing, and reference-guided editing, which demonstrates 10.96$\times$ acceleration than the current state-of-the-art editing approach.
\end{itemize}

\begin{figure*}[htbp]
    \centering
    \includegraphics[width=\textwidth]{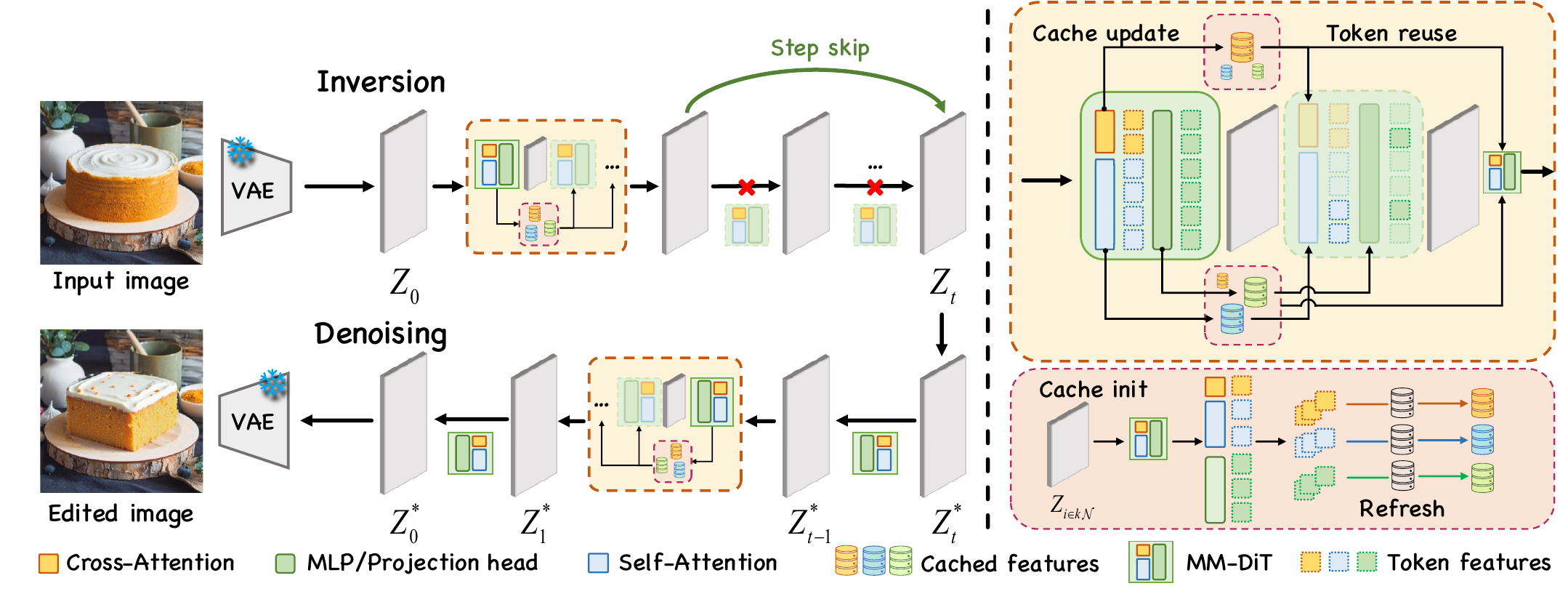}
    \caption{\textbf{The overview of our approach}. The proposed framework for image editing based on MM-DiT diffusion models employs an efficient denoising and training-free approach. The pipeline takes the original image and an editing prompt as inputs. Specifically, the cache is refreshed entirely in fixed time-step interval, while partial computation for updating cache is maintained for the intermediate timesteps.}
    \label{fig:framework}
\end{figure*}

%% file: sec/2_related.tex
\section{Related Work}
\subsection{Acceleration of Diffusion Models}
Low-latency, high-quality generation methods are an important research direction. Currently, there are two main types of diffusion model acceleration methods: 
the first one is to reduce the number of sampling steps~\cite{Song2022DDIM,lu2022dpmsolverfastodesolver,lu2023dpmsolverfastsolverguided,liu2022flowstraightfastlearning}, and the second one is to accelerate internal computation of the diffusion model.
Solutions to reduce computational complexity include model distillation and compression~\cite{Xie2024SANA}, token pruning~\cite{zhang2025sito}, token merging~\cite{bolya2023tokenmergingvitfaster,Feng2024DiT4Edit, bolya2023tomesd}, and layer-wise caching techniques~\cite{deepcache,gao2024dual,L2C,Liu2024FasterDiffusion,fora}.
However, layer-wise caching techniques have a large cache granularity, which ignore the asymmetry of importance at the token level. Toca~\cite{toca} and Duca~\cite{duca} adopt a token-wise cache, focusing on and assigning importance to tokens with score maps.
During recomputation, a certain proportion of tokens are selected for refreshing, achieving a lossless acceleration. 

Unfortunately, existing caching schemes are not specifically designed to accommodate the characteristics of editing tasks and suffer from the following issues:

First, computing intrinsic feature correlations between tokens introduces additional internal overhead in caching operations. Second, some caching strategies and editing methods require accessing attention maps~\cite{toca} or storing KV matrices~\cite{tf-icon,zhu2025kv}, leading to incompatibility with existing Transformer acceleration techniques, such as FlashAttention~\cite{flashattention,yang2025videograin}, thereby increasing latency. Moreover, editing tasks inherently exhibit prior information regarding the spatial distribution of regions requiring greater attention, yet current caching techniques fail to leverage it.

Our approach significantly improves acceleration over traditional caching methods by effectively reducing both temporal and spatial redundancy. Additionally, we introduce token index preprocessing to further compress the extra overhead incurred by caching. As a result, our method simultaneously optimizes both editing performance and acceleration efficiency among various editing tasks.
\subsection{Image Editing}
Image editing, as an important application in the generative field, has received widespread research and exploration in the academic community. This includes controllable text-to-image generation, image inpainting, and image-to-image generation, among other approaches~\cite{ma2022visual,ma2023magicstick,ma2024followyourclick,ma2024followyouremoji,ma2024followyourpose,wang2024cove,Wang2024TamingRectifiedFlow,kulikov2024flowedit}.
Common editing approaches follow a noise addition and denoising framework, where the original image is perturbed by a certain level of noise in the latent space to leverage the model's editing capabilities. The final edited image is then obtained through a denoising process that restores clarity.
Training-free inversion editing techniques, when applied to editing tasks such as prompt-guided editing, image composition, and image dragging, involve operations on the attention map, including modification, enhancement, and replacement. These methods have been applied in P2P~\cite{p2p} and more subsequent works~\cite{tf-icon,zhu2025kv,meng2024anidoc,guo2024refir}.
Since inversion-based approaches require the inversion technique to reconstruct the denoised image and significantly impact editing quality, researchers have explored both training-based and training-free inversion techniques. InfEdit adopts a virtual inversion strategy without explicit inversion during sampling, enabling accurate and consistent editing, and falls under the category of inversion-free image editing methods.
Unfortunately, these methods based on or utilizing attention maps require storing the attention map and manipulating the corresponding KV matrices~\cite{p2p,zhu2025kv}. This results in incompatibility with common attention acceleration techniques, increasing inference latency. In contrast, our method enables reduces the redundancy in inversion and denoising, enhancing the efficiency and speed of editing.

%% file: sec/3_preliminaries.tex
\section{Preliminaries}
\subsection{Rectified Flow}
The Rectified Flow~\cite{flowmatch} method models the transformation from a Gaussian noise distribution \( \pi_0 \) to the real data distribution \( \pi_1 \) as a continuous change along a straight path by learning a forward simulation system.


In the forward process, the state of the system can be viewed as a linear interpolation between the initial state and the Gaussian noise, which is simplified and easier to understand compared to traditional methods~\cite{ddpm}. 
\[
\mathbf{X}_t = (1 - t)\mathbf{X}_1 + t\mathbf{X}_0, \quad X_1\sim\pi_1, X_0 \sim \pi_0 \tag{1}
\]

By differentiating the above expression with respect to \( t \), the ODE form of the Rectified Flow can be obtained: \(\frac{d\mathbf{X}_t}{dt} = \mathbf{X}_1 - \mathbf{X}_0\).
Furthermore, let us define \( v_\theta = \frac{dX_t}{dt} \), the training objective can then be transformed into minimizing the integral of this expectation over the time steps:
\[
\min_{\theta} \int_0^1 \mathbb{E} \left[\left\|\mathbf{X}_1 - \mathbf{X}_0 - v_\theta(\mathbf{X}_t, t)\right\|^2\right] dt\tag{2}
\]
Here, \( \theta \) represents the parameters of the neural network to be optimized. Due to the complexity introduced by the integral symbol, the optimization of these parameters is typically performed using the equivalent form of Conditional Flow Matching (CFM):
\[
\mathcal{L}_{CFM} = \mathbb{E}_{t, p_t(z|\epsilon), p(\epsilon)} \left\| v_{\Theta}(z,t) - u_t(z|\epsilon) \right\|_2^2, \tag{3}
\]
where the conditional vector fields \( u_t(z|\epsilon) \) provides an equivalent yet tractable objective.
 \( p_t \) is the probability path between \( p_0 \) and \( p_1 \), and \( p_1 \sim \pi_0 \).

%% file: sec/4_approach.tex
\section{Approaches}
Our approaches aim to reduce the spatial and temporal redundancy to improve the efficiency of image editing. The core idea is to leverage the mask of edited area to guide 
cache refreshing and reuse.
The pipeline of the EEdit is shown in Figure.~\ref{fig:framework}. In this section, we will describe the design of spatial locality caching~(see Section.~\ref{sec:Spatial Locality Caching}) and token index preprocessing~(see Section.~\ref{sec:Token Index Preprocessing}) for spatial redundancy. For temporal redundancy, we introduce the inversion step skipping in the Section.~\ref{sec:inversion step skipping}.

\begin{algorithm}[ht]
\caption{\textbf{SLoC} Editing with \textbf{ISS}\&\textbf{TIP}}
\label{alg:edit}
\begin{algorithmic}[1]
\Require Input image \(\mathbf{I_s}\), Mask for editing region \(\mathbf{M_s}\), Prompt for editing \(\mathbf{P_m}\), Randomly initialized map \(\mathcal{R}\), Bonus for edited region \(\mathbf{S_E}\), Cache dict \(\mathcal{C}_t[l,f]\) and NN layer \(\mathcal{F}_i\). Token-wise multiplication \(\odot\) and addition \(\oplus\).
\Ensure The edited result $\mathbf{I}^*$
\State // \textbf{Token Index Preprocessing}
\State $\mathcal{M}_{freq} \gets \boldsymbol{zero}\mathbf{[\mathcal{F}_i,l]}$
\For{$t = T,T-1, \dots 1$}
    \For{$\mathcal{F}_i \gets {\mathbf{{SA}}_l,\mathbf{CA}_l,\mathbf{MLP}_l},l \in [1,2\dots\mathcal{L}]$}
    \State \(\mathcal{S}_l \gets  (\mathcal{R} \odot \mathbf{S_E})\oplus \mathcal{M}_{freq} \)
    \State \(\mathcal{I}_{i,l,t} \gets \mathbf{Sel}_{topR\%}(\mathcal{S}_l)\)
    \State\(\mathbf{Update}(\mathcal{M}_{freq})\)
    \EndFor
\EndFor
\State $\mathbf{Z}_0 \gets \mathbf{cat}(\text{VQ-Encoder}(\mathbf{I_s}),\text{Txt-Encoder}(\mathbf{P_m}))$
\State // \textbf{Inversion Reduction}
\For{$t = 1,2 \dots, T$}
    \State $\mathbf{Z}_t \gets 
    \begin{cases} 
        \mathbf{Z}_{t-1}  \quad\mathbf{if} \;
         t \bmod m \neq 1 \;\mathbf{and} \; m \neq T ,m \in \mathcal{N}
        \\
        \text{RF-inversion}(\mathbf{Z}_{t-1}, t-1, \mathcal{\phi})  \quad \mathbf{otherwise}
    \end{cases}$
\EndFor
\State \(\mathbf{Z}_{T}^{*}\gets \mathbf{Z}_T\)
\State // \textbf{Image Editing Steps with SLoC}
\For{$t = T,T-1, \dots 1$}
    \For{$\mathcal{F}_i \gets {\mathbf{SA}_l,\mathbf{CA}_l,\mathbf{MLP}_l},l \in [1,\dots\mathcal{L}]$}
    \State \(Z_{l+1}^*\gets \mathbf{scatter}(\mathcal{F}_i(Z_{l}^*,\mathcal{I}_{i,l,t}),\mathcal{C}_{t+1}[l,\mathcal{F}_i])\)
    \State\(\mathbf{Update}(\mathcal{C}_{t+1}[l,\mathcal{F}_i])\)
    \EndFor
    \State $\mathbf{Z}_{t-1}^* \gets \mathbf{Z}_{t-1}^{*} \odot \mathbf{M}_s + \mathbf{Z}_{t} \odot (1 - \mathbf{M}_s)$
\EndFor
\State $\mathbf{I}^* \gets \text{VQ-Decoder}(\mathbf{Z}_0^*)$
\State \Return $\mathbf{I}^*$
\end{algorithmic}
\end{algorithm}

\begin{figure*}[ht]
    \centering
    \includegraphics[width=0.87\linewidth]{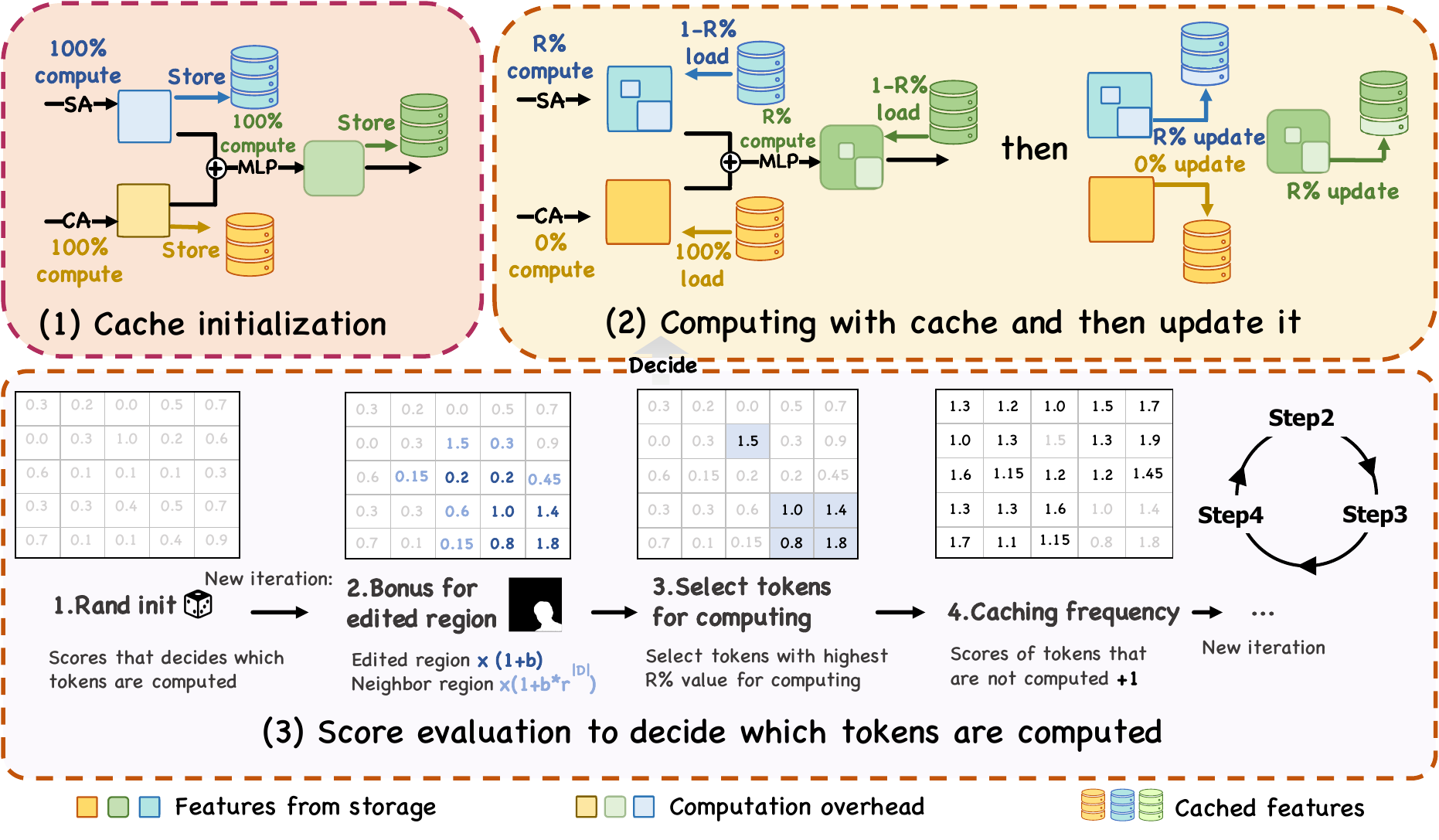}
    \caption{
    \textbf{The pipeline of spatial locality caching.}
    (1) The initialization and refresh process of cache storage using the computed results from SA (Self-Attention), CA (Cross-Attention), and MLP.  
    (2) The token-wise partial computation logic and the cache update mechanism.  
    (3) The initialization and update logic for scoring, which is responsible for selecting indices for partial computation.
    }
    \label{fig:methods}
\end{figure*}
\subsection{Spatial Locality Caching}
\label{sec:Spatial Locality Caching}
\noindent\textbf{Score Bonus for Editing Region.}

In mask-guided image editing tasks, the tokens corresponding to image patches outside the edited region are actually derived from \( Z_T^{inv} \), which is obtained from inversion process. This part is directly replaced in the latent space.

As a result, the importance of DiT's computation on these tokens is significantly reduced. Instead, we aim to focus more on the computation of tokens within the edited region. Simultaneously, to account for the modeling of correlations between important tokens and their neighboring tokens, the neighboring tokens are rewarded with gradually decreasing importance. We designed a score bonus map \(\mathbf{S_E}\) to control the reward intensity, defined as:
\[
\mathbf{S_E}(\mathbf{x}) =
\begin{cases} 
1 + b \cdot r^k, &  \mathbf{x} \in \mathcal{N}_k(M_s), \; k\in0, 1, \ldots, K \\
1, &  \mathbf{x} \notin \bigcup_{k=0}^{K} \mathcal{N}_k(M_s)
\end{cases}
\]Here \(\mathbf{x}\) represents an arbitrary token in DiT. \(M_s\) is the latent mask for point set of edit region. Hyperparameters such as \( b > 1 \) represent the bonus factor, while \( 0 < r < 1 \) denotes the decay ratio. \(K\) is the border‐score smoothing hyperparameter. The neighborhood of \( M_s \) with an L1 norm equal to \( k \) is denoted as \( \mathcal{N}_k(M_s) \) and is defined as:
\[
\mathcal{N}_k(M_s) = \{ \mathbf{x} : \exists \mathbf{e} \in M_s, \|\mathbf{x} - \mathbf{e}\|_1 = k \}.
\]
\noindent\textbf{Update Strategy with Cache Frequency Control.}
When the proportion of tokens corresponding to the edited region is lower than the full computation of the network, meaning it is smaller than the cache refresh ratio, not all tokens in the edited region undergo the same level of full computation. Instead, we prioritize recomputing and refreshing in the cache those tokens that have been updated less frequently and reused more times.

SLoC track the frequency of times each token is reused from the cache and increment its score accordingly on a token-wise map \(\mathcal{M}_{freq}\). The more frequently a token is reused, the more likely its features will be recomputed and refreshed. Once certain tokens undergo recomputation and refresh, their corresponding usage frequency counters are immediately reset to zero. 
From another perspective, the design of cache frequency control serves two key purposes. First, it encourages the recomputation of frequently reused tokens to reduce accumulated errors. Second, it suppresses the redundant recomputation of tokens that are repeatedly updated, thereby reducing computational overhead.

\subsection{Token Index Preprocessing}
\label{sec:Token Index Preprocessing}

Although SLoC reduces spatial redundancy  by using caching, as a double-edged sword, such an acceleration comes with certain limitations: in a cache cycle shown in Figure.~\ref{fig:methods}, the cache overhead is sequentially arranged as follows: (1)  Randomly initialize. (2) Bonus for edited region. (3) Perform sorting and selection. (4) Update. (5) Refresh. These steps constitute the caching cycle and operate as illustrated in Figure.~\ref{fig:methods}.

However, we observe that internal score updates and token selection take additional computational costs in editing processing and can be further optimized.

Our initialization and update strategy for the score map can be transformed from an online operation into an offline algorithm while maintaining full mathematical equivalence~(See the supplementary material for the formal proof). 

The key insight here is that under the score update rule

\begin{equation*}
   \mathcal{S} \gets (\mathcal{R} \odot \mathbf{S_E})\oplus \mathcal{M}_{freq},
\end{equation*}
we can prove the top-\( R\% \) selected indices \( \mathcal{I}^{(t)}_{\text{topR\%}}\) in the same time step in the offline and online process remain equivalent:
\begin{equation*}
   \mathcal{I}^{(t)}_{\text{topR\%}}(\text{offline}) = \mathcal{I}^{(t)}_{\text{topR\%}}(\text{online}) \quad \forall t\in [1 \dots T].
\end{equation*}
Here \(\mathcal{S}\) denotes the score map, \(\mathcal{R}\) represents the randomly initialized score values, \(\mathbf{S_E}\) corresponds to the region score bonus, and \(\mathcal{M}_{freq}\) denotes the matrix for cache frequency control, consistent with Algorithm~\ref{alg:edit}.

Therefore, we can decouple the cache score update, sorting, and index selection logic from the model's computation process. By precomputing and storing the required token indices during preprocessing, the overhead from cache operations in the inference phase is reduced to a single read/write cost. Consequently, the former 4 steps of the cache cycle can be omitted during inference.
SLoC directly updates only the necessary tokens during inference, without executing any redundant score computation or update strategies.

\subsection{Inversion Step Skipping}
\label{sec:inversion step skipping}

After addressing the spatial redundancy, we focus on the temporal redundancy in both inversion/denoising processing. As demonstrated in Fig.~\ref{fig:qualitative-intro}(b), previous methods~\cite{p2p, rfinversion} calculate the noise at each timestep. However, the emergence of DDIM~\cite{Song2022DDIM} motivates us to consider \textit{whether we can skip certain steps in flow matching-based inversion}.  During the editing process, we discover that skipping some inversion steps does not degrade the quality of the editing results(see Fig.~\ref{fig:qualitative-intro}(b)), which demonstrates the redundancy of the inversion process. 

To eliminate this redundancy, we propose the inversion step skipping strategy in the editing processing. The key insight of inversion step skipping is that \textit{skipping some inversion steps does not result in noticeable artifacts, but it can significantly improve speed}. Formally, during the inversion, the skipping step is set $m$. We add the noise from the $\mathbf{Z}_{0}$. In every 
$m$ steps, we perform an rf-inversion~\cite{rfinversion}, while for the other timesteps, we directly reuse the noise from the previous timestep. This process can be formally expressed as follows:

$\mathbf{Z}_t \gets 
    \begin{cases} 
        \mathbf{Z}_{t-1}  \quad\mathbf{if} \;
         t \bmod m \neq 1 \;\mathbf{and} \; m \neq T ,m \in \mathcal{N}
        \\
        \text{RF-inversion}(\mathbf{Z}_{t-1}, t-1, \mathcal{\phi})  \quad \mathbf{otherwise}
    \end{cases}$
Additionally, we provide the pseudocode implementation in Algorithm.~\ref{alg:edit}. Note that, to ensure noise quality, the final step of the inversion process is always fully computed, whereas the intermediate steps employ a mixed strategy of full computation and cache-accelerated computation. In our experiment, skipping step $m$ is set 3 to balance the quality and speed. We also provide the ablation study about the inversion skipping step in Section.~\ref{sec:ablation2}.

%% file: sec/5_exp.tex
\section{Experiments}
\begin{figure*}
    \centering
    \includegraphics[width=\linewidth]{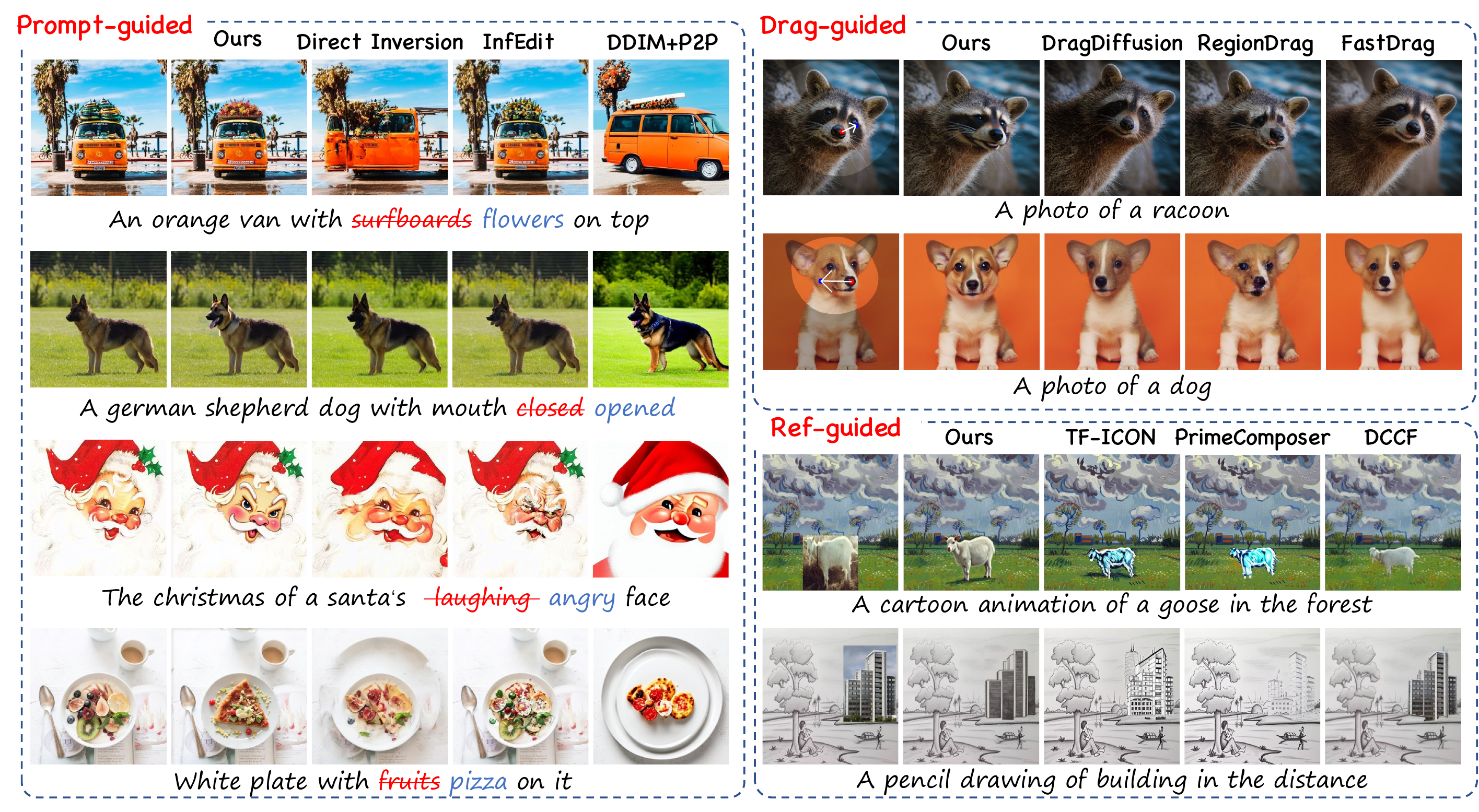}
    \caption{\textbf{Qualitative comparisons between baselines and our approach}. We compare our approach with  various  editing tasks, including prompt-guided image editing, drag-guided image editing, and reference-guided image editing}
    \label{fig:qualitative}
\end{figure*}

\subsection{Implementation Details}
In our experiment, the base model adopted the currently popular flow matching model, FLUX-Dev~\cite{FLUX1Dev2024}, which consists of 12B parameters. Models for other qualitative results are implemented using SD series~\cite{StableDiffusionV14,LCMDreamshaperV7,Rombach_2022_CVPR,StableDiffusion21Base} with original codebase.  Our inference and editing pipeline is built upon the codebase from Hugging Face. The inversion and denosing step is set to 28. RF-inversion relative hyper-parameters follow original implementations~\cite{rfinversion}.  We employ the text-guidance ratio of 7.0.  
We selected the PIE-Bench~\cite{PIE-Bench2024} Benchmark as the dataset for prompt-guided editing, the TF-ICON Test Benchmark~\cite{lu2023tf} as the dataset for reference-guided editing, and DragBench-SR and DragBench-DR as the datasets~\cite{RegionDrag2024} for drag-guided editing.
All experiments were conducted on an NVIDIA H20. More details and evaluation metrics can be found in supplementary material.
\subsection{Compare with Baseline}
\noindent\textbf{Qualitative Comparison.}
We perform  extensive qualitative comparisons between our method and various editing approaches. 
Specifically, we evaluate three kinds of popular editing tasks,
including prompt-guided, drag-guided, and reference-guided image editing.
(1) \textbf{Prompt-guided image editing}: we compare with Direct Inversion ~\cite{pnp}, InfEdit~\cite{Xu2023InversionFreeEditing}, and P2P~\cite{p2p}.
(2) \textbf{Drag-guided image editing}, we 
select three methods to do comparisons, including DragDiffusion~\cite{shi2023dragdiffusion}, Region Drag\cite{regiondrag}, and FastDrag~\cite{zhao2024fastdrag}. Pipeline are implemented from public codebase from github.
(3) \textbf{Reference-guided editing}, we compare our approach with TF-ICON~\cite{tf-icon}, PrimeComposer~\cite{Wang2024PrimeComposer}, and DCCF~\cite{dccf}. SD-v2-1 checkpoint is used for Image Composition.
As shown in Figure.~\ref{fig:qualitative},
for prompt-guided task, Direct inversion and InfEdit fail to edit the image successfluly. DDIM$+$P2P has a challenge to maintain the background consistency. By contrast, our approach exhibits superior consistency, enhanced detail preservation, and improved aesthetic quality in terms of style.

\begin{table}[ht]
    \centering
    \caption{\textbf{Comparison of quality and efficiency}}
    \label{tab:comparison_others}
    \scalebox{0.49}{
    \begin{tabular}{llccccccc}
    \toprule
    \multirow{2}{*}{\textbf{Method}} 
      & \multirow{2}{*}{\textbf{Inversion}} 
      & \multicolumn{4}{c}{\textbf{Quality metrics}} 
      & \multicolumn{3}{c}{\textbf{Efficiency metrics}} \\
    \cmidrule(lr){3-6} \cmidrule(lr){7-9}
      & & PSNR$\uparrow$ 
          & LPIPS$_{\times10^{-2}}$\,$\downarrow$ 
          & SSIM$\uparrow$ 
          & CLIP-T$\uparrow$ 
          & FLOPs(T) 
          & Time (s) 
          & Model type \\
    \midrule
    P2P
      & DDIM 
      & 17.87 & 20.88 & 0.72 & 25.13 
      & 334.4 & 18.75 & SD1-4 860M\\
    MasaCtrl
      & DDIM 
      & 22.19 & 10.54 & 0.80 & 24.02 
      & 334.4 & 17.65 & SD1-4 860M\\
    MasaCtrl
      & DI 
      & 22.69 &  8.73 & 0.82 & 24.39 
      & 408.3 & 23.60 & SD1-4 860M \\
    PnP
      & DDIM 
      & 25.23 & 11.27 & 0.80 & 25.42 
      & 180.9 & 13.81 & SD1-5 860M \\
    PnP
      & DI 
      & 22.46 & 10.55 & 0.80 & 25.48 
      & 180.9 & 13.77 & SD1-5 860M\\
    InfEdit
      & Inversion-free 
      & \textbf{28.11} &  \textbf{5.61} & \textbf{0.85} & \textbf{25.86} 
      & \textbf{124.6}  &  \textbf{2.90} & SD1-5, LCM 860M \\
    \midrule
    DiT4Edit
      & DPM-Solver++ 
      & \textbf{22.85} &    -- &   -- & \textbf{25.39} 
      & \textbf{260.4}  &  \textbf{5.15} & PixArt XL 610M \\
    \midrule
    RF-inv 
      & RF-inversion
      & 17.74 & 24.40 & 0.66 & 26.31 
      & 1111.6 &    13.56 & FLUX 12B \\
    RF-Edit 
      & RF-Solver 
      & 20.17 & 18.50 & 0.77 & \textbf{26.64} 
      & 2223.2 &   26.23 & FLUX 12B \\
    Flow-Edit 
      & -- 
      & 22.20 & 10.49 & 0.85 & 25.80
      &  952.8 &    11.84 & FLUX 12B \\
    \rowcolor{green!8}
    SLoC 
      & RF-inversion 
      & \textbf{31.97} &  1.96 & \textbf{0.94} & 25.37 
      &  384.0 &    5.96 & FLUX 12B \\
    \rowcolor{green!8}
    SLoC 
      & ISS 
      & \textbf{31.97} & \textbf{1.95} & \textbf{0.94} & 25.38 
      &   \textbf{264.5}   &  \textbf{4.60} & FLUX 12B \\
    \bottomrule
    \end{tabular}
    }
\end{table}

    

\noindent\textbf{Quantitative Comparison.} We compared various editing methods in terms of background consistency in non-edited regions and adherence to prompts in the edited regions. As shown in Table.~\ref{tab:comparison_others}, since our approach allocates fewer computational resources to non-edited regions and employs masking in the latent space, we achieve significantly better performance in background consistency compared to other state-of-the-art editing methods. Furthermore, in terms of prompt adherence, the Inversion step skipping strategy achieves a CLIP score comparable to that of RF-Inversion with full computation. It is also competitive with others.

\subsection{Ablation Study}
Extensive quantitative and qualitative ablation experiments were conducted to analyze the impact of our methods on editing quality. We measure the similarity to the input image in non-edited regions as \textit{background preservation (BG preservation)}. Additionally, we compare the editing results incorporating various modules with the editing results obtained from full computation without caching as \textit{foreground fidelity (FG fidelity)}.

\noindent\textbf{Ablation study of spatial locality caching.}
\begin{figure}
    \centering
    \includegraphics[width=\linewidth]{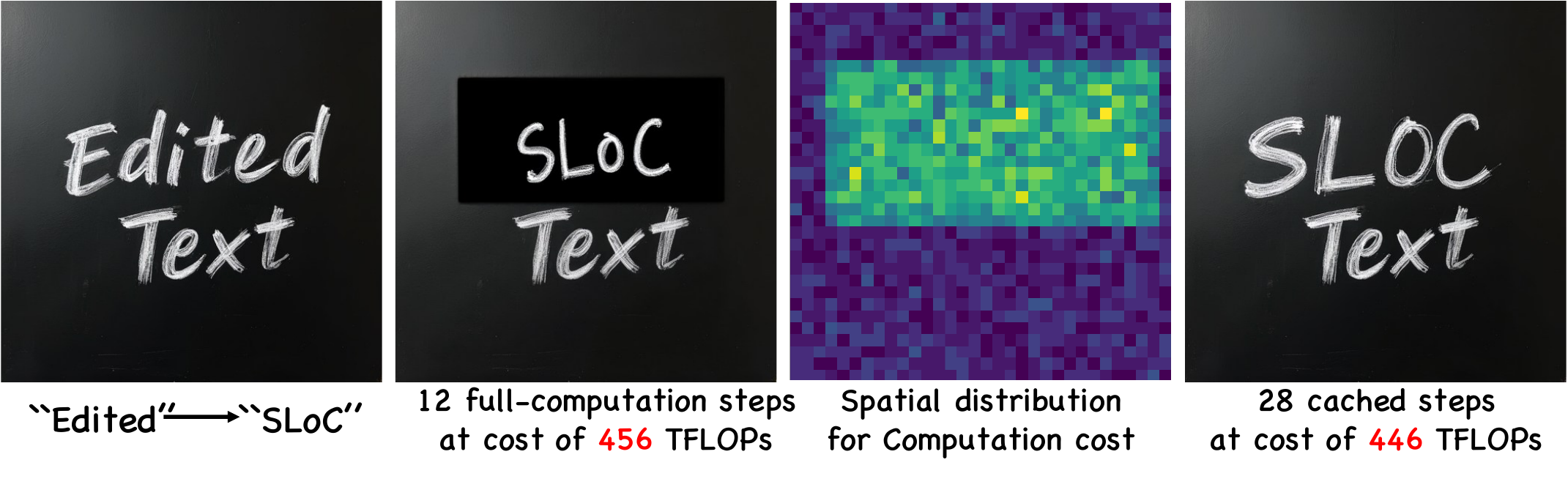}
    \caption{\textbf{A qualitative example of ablation over SLoC}, demonstrating that regional computation with SLoC leads to higher quality and lower computational overhead.
    }
    \label{fig:ablation_sloc}
\end{figure}
Figure.~\ref{fig:ablation_sloc} illustrates the impact of SLoC on editing performance. In text editing tasks, SLoC leverages the prior of spatial locality to achieve superior editing results compared to the original full-computation editing method, despite using more steps but incurring lower computational overhead. This validates the feasibility of our regionally focused computation strategy.
Furthermore, quantitative comparisons with other cache-based acceleration methods demonstrate the superiority of our approach (shown in supplementary material). 

\begin{table}[ht]
    \centering
    \caption{\textbf{Ablation study on ISS}. Ablation study is conducted on different skipping settings for background preservation, foreground fidelity and inference time.}
    \label{tab:ablation_inversion}
    \scalebox{0.6}{
        \begin{tabular}{cc|c|ccc|c}
            \toprule
            \multirow{3}{*}{\textbf{Inversion}} & 
            \multirow{3}{*}{\textbf{Denoising}}  & 
            \multicolumn{1}{c|}{\textbf{BG preservation}} & 
            \multicolumn{3}{c|}{\textbf{FG fidelity}} & 
            \multirow{2}{*}{\textbf{Inference} $\downarrow$} \\
            \cmidrule(lr){3-3} \cmidrule(lr){4-6}
            & & \textbf{LPIPS} $_{\times 10^{-2}}^\downarrow$ & \textbf{LPIPS} $_{\times10^{-3}}^\downarrow$ & \textbf{PSNR $\uparrow$} & \textbf{FID$\downarrow$} &  Time (s) \\
            \bottomrule
            \midrule
            \textbf{Full step} & \textbf{Full step} & 1.98 & - & - & - & 13.27 \\
            \midrule
            2-step skip & \multirow{3}{*}{\textbf{Full step}} & 1.98 & 5.46 &  43.77& 3.35 & 10.16 \\
            3-step skip&  & 1.98 & 5.29 &  43.99& 3.23 & 9.31  \\
            4-step skip&  & 1.98 & 5.29 & 43.80 & 3.31 & 8.76  \\

            \bottomrule
        \end{tabular}
    }
\end{table}

\begin{figure}
    \centering
    \includegraphics[width=1.0\linewidth]{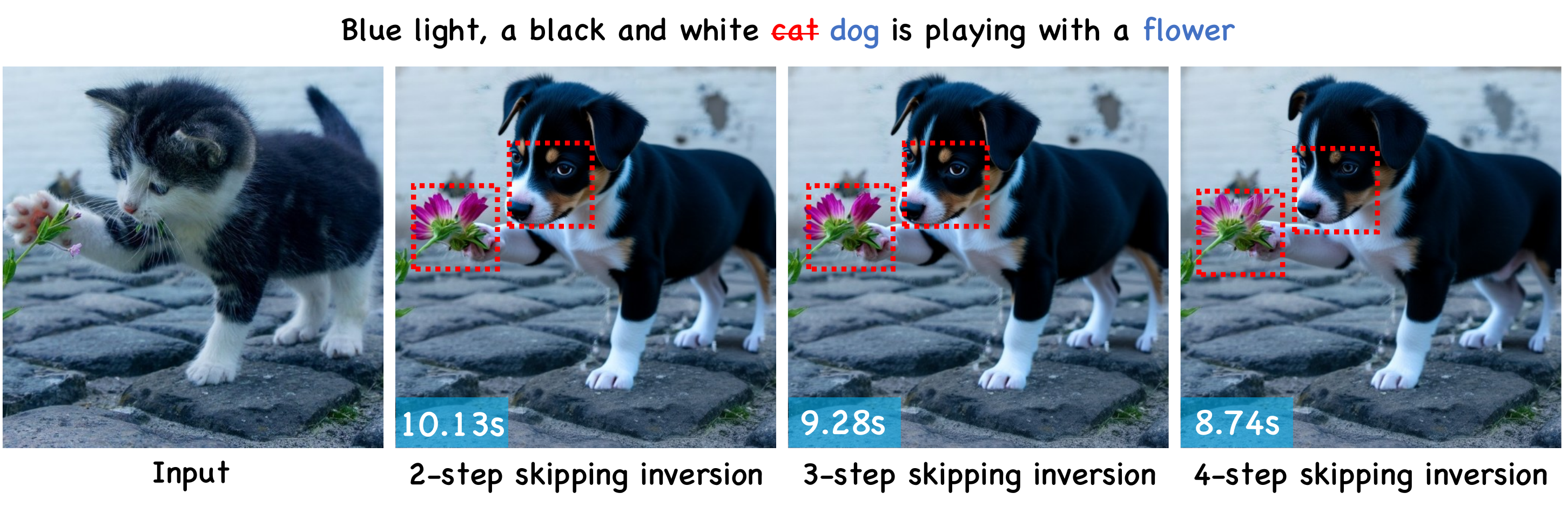}
    \caption{\textbf{The qualitative ablation study about inversion step skipping. } We visualize results in various inversion step skipping.}
    \label{fig:ablation_inversion}
    
\end{figure}
\begin{table}[th]
    \centering
    \caption{
    \textbf{Ablation study on different configurations of TIP and ISS} in  prompt-guided, drag-guided, and ref-guided editing.}
    \label{tab:ablation_fig}
    \scalebox{0.6}{
    \renewcommand{\arraystretch}{1.2}
    \setlength{\tabcolsep}{6pt} 
        \begin{tabular}{c|ll|ccc|cc}
            \toprule
            \multirow{2}{*}{\textbf{TASK}} & \multicolumn{2}{c}{\textbf{Method}} & \multicolumn{3}{|c|}{\textbf{FG fidelity}} & \multirow{2}{*}{\textbf{Inference (s) $\downarrow$}} & \multirow{2}{*}{\textbf{CLIP-E} $\uparrow$} \\
            \cmidrule(lr){2-6}
            & \textbf{TIP} & \textbf{ISS} & FID\(\downarrow\) & PSNR\(\uparrow\) &LPIPS$^{\downarrow}_{\times 10^{-2}}$ &  &  \\
            \midrule
            \multirow{4}{*}{\textbf{Prompt}} 
            & \texttimes & \texttimes & 39.50 & 31.75 & 5.75 & 5.96 & 21.34 \\
            & \texttimes & \checkmark & 39.33 & 31.76 & 5.74 & 5.06 & 21.34 \\
            & \checkmark & \texttimes & 39.42 & 31.75 & 5.75 & 5.14 & 21.34 \\
            & \checkmark & \checkmark & \textbf{39.21} & \textbf{31.76} & \textbf{5.74} & \textbf{4.60} & \textbf{21.34} \\
            \midrule
            \multirow{2}{*}{\textbf{Dragging}} & \texttimes & \texttimes & \textbf{20.61} & 33.47 & 2.28 & 7.12 & 22.20 \\
            & \checkmark & \checkmark &  22.07 & \textbf{33.68} & \textbf{2.19} & \textbf{5.66} & \textbf{22.21} \\
            \midrule
            \multirow{2}{*}{\textbf{Composition}} & \texttimes & \texttimes & \textbf{12.33} & 39.78 & 0.54 & 7.25 & 23.35 \\
            & \checkmark & \checkmark & 12.35 & \textbf{39.80} & \textbf{0.54} & \textbf{5.66} & \textbf{23.35} \\
            \bottomrule
        \end{tabular}
    }
    
\end{table}
\begin{figure}[ht]
    \centering
    \includegraphics[width=0.9\linewidth]{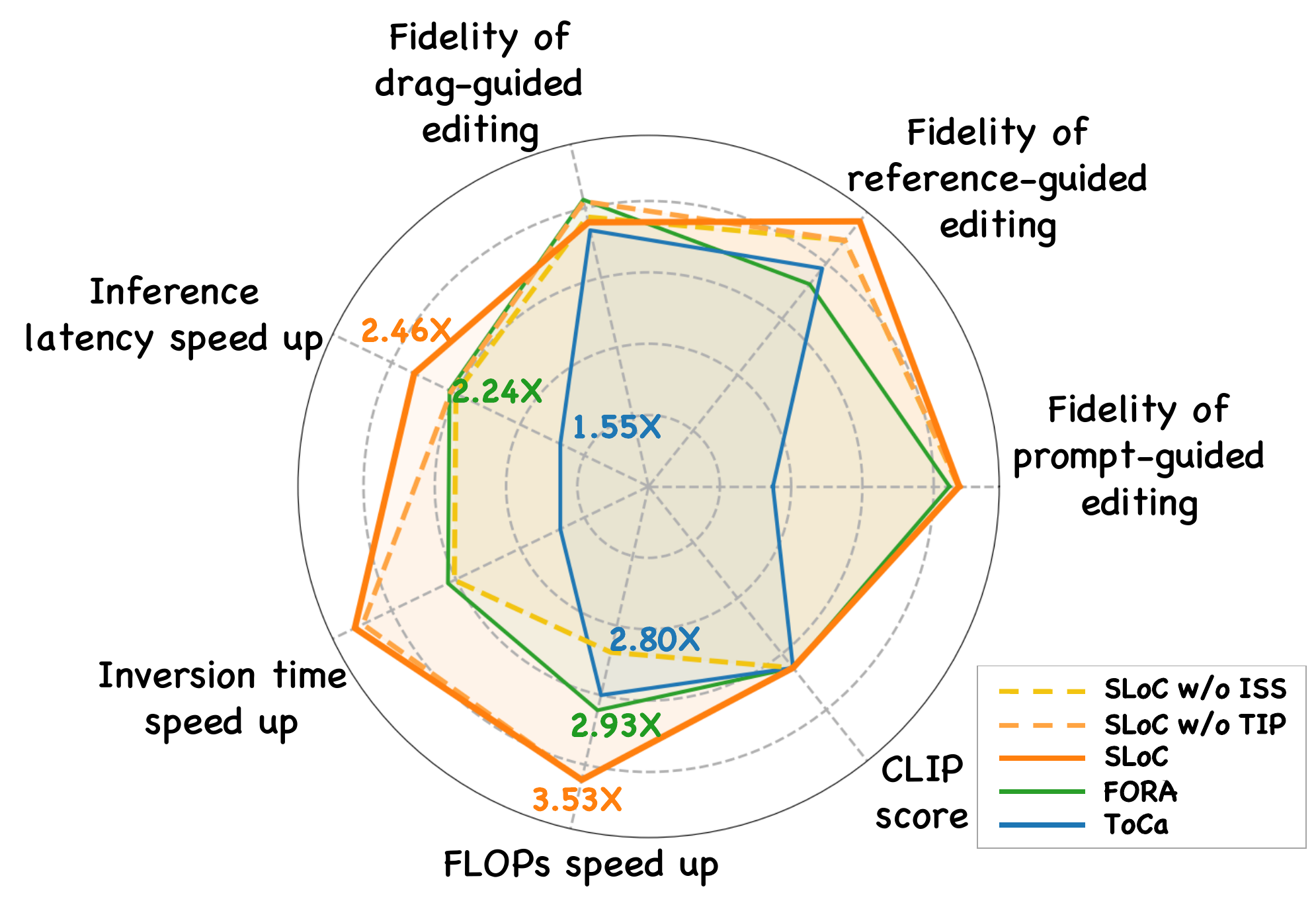}
    \caption{\textbf{A Comparison over different configurations and other cache methods.} Fidelity of various tasks, speed up ratio and clip score are shown in this radar chart. }
    \label{fig:ablation_fig}
\end{figure}

\noindent\textbf{Ablation study of inversion step skipping.}
It can be observed in Table.~\ref{tab:ablation_inversion} and Figure.~\ref{fig:ablation_inversion} that increasing the skip interval \( m \) to reduce inversion computation significantly decreases inference time while maintaining background preservation without quality degradation. Additionally, the foreground fidelity metrics exhibit negligible differences compared to full-step computation. This validates the effectiveness and fidelity of our ISS strategy.

\noindent\textbf{Ablation study of TIP over ISS and various tasks.}
\label{sec:ablation2}
As shown in Table~\ref{tab:ablation_fig} and Figure~\ref{fig:ablation_fig}, TIP remains compatible with ISS while maintaining comparable FG fidelity and CLIP scores across various editing tasks. Furthermore, it achieves an additional reduction in inference latency by an average of over 20\% compared to the SLoC baseline.

%% file: sec/6_conclusion.tex
\section{Conclusions}
Inversion-based image editing usually suffers from expensive computation costs caused by the spatial and temporal redundancy within diffusion models. To solve this problem, we design an efficient editing framework using the spatial locality caching with token index preprocessing  and inversion step skipping.  
To the best of our knowledge, we are the first to adapt cache-based acceleration for diffusion inference across various editing tasks. Our work provides valuable insights and exploration for future approaches toward efficient, and potentially real-time, image and even video editing.

\noindent \textbf{Acknowledgment} This work was partially supported by Dream Set Off - Kunpeng\&Ascend Seed Program and The Research Grant Council of the Hong Kong Special Administrative Region under grant number 16203122. 

%% file: sec/appendix.tex
\clearpage
\setcounter{page}{1}
\maketitlesupplementary

\definecolor{mygray}{gray}{.92}
\definecolor{ForestGreen}{RGB}{34,139,34}
\newcommand{\fg}[1]{\mathbf{\textcolor{ForestGreen}{#1}}} 

\definecolor{Forestred}{RGB}{220,50,50}
\newcommand{\fr}[1]{\mathbf{\textcolor{Forestred}{#1}}}  

\section{Analysis on Hidden States in MM-DiT}

\begin{figure}[ht]
    \centering
    \includegraphics[width=0.7\linewidth]{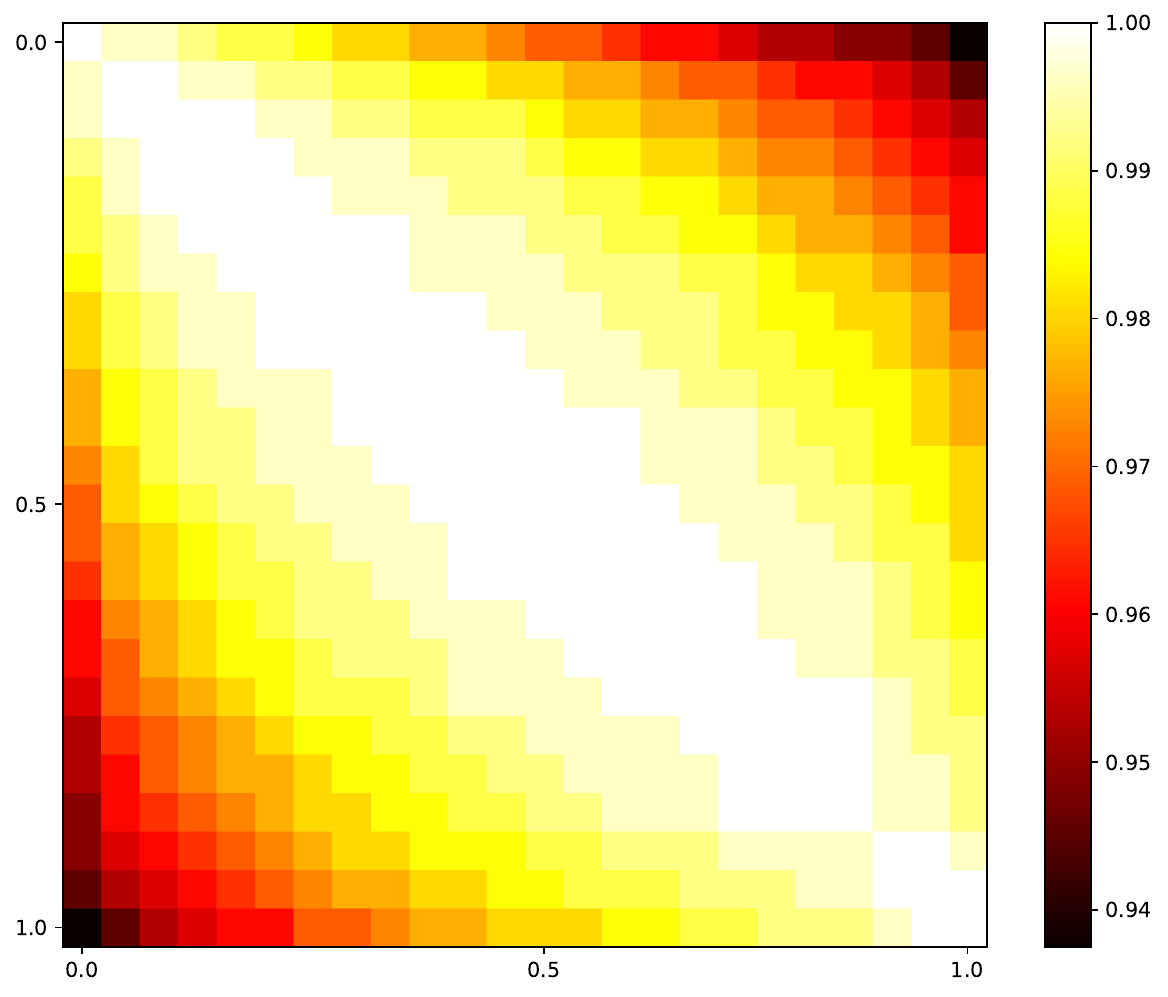}
    \caption{Cross-Attention hidden states similarity}
    \label{fig:hiddenstates1}
    \includegraphics[width=0.7\linewidth]{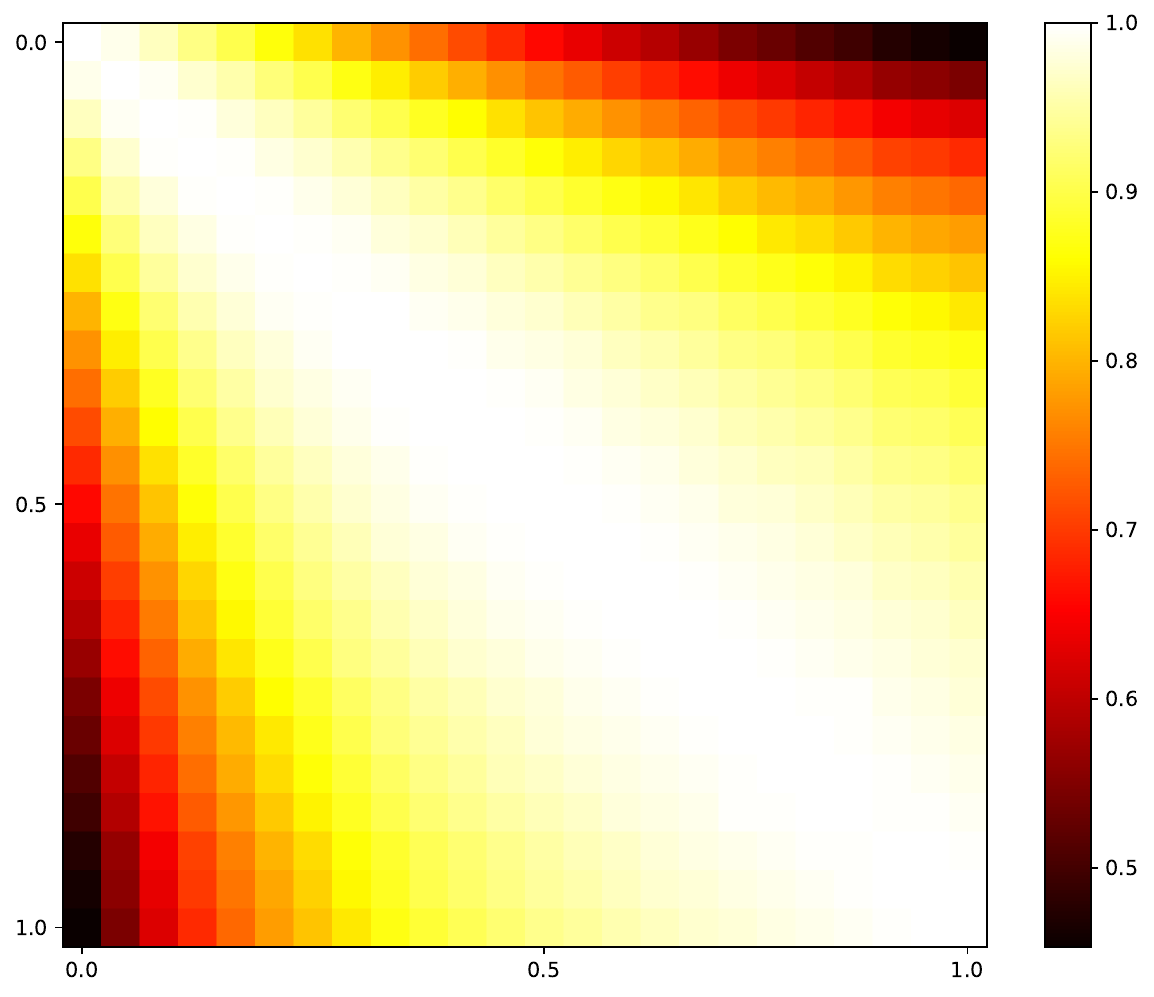}
    \caption{Self-Attention hidden states similarity}
    \label{fig:hiddenstates2}
\end{figure}
As shown in the Figure.~\ref{fig:hiddenstates1} and Figure.~\ref{fig:hiddenstates2}, we visualize the cosine similarity of hidden states across different timesteps in both Cross-Attention and Self-Attention. It can be observed that Cross-Attention exhibits higher similarity, indicating greater redundancy in this module. Consequently, in our approach, Cross-Attention is either fully computed or entirely skipped to optimize efficiency.

\section{Algorithm for Image Editing with SLoC}
\subsection{Cache Frequency Control}
Cache Frequency Control can be formulated as

\begin{equation*}
\mathcal{S}^{\tau+1}_{ij} \leftarrow 
\begin{cases} 
\mathcal{S}_{ij}^{\tau} + \gamma f_{ij}^{\tau} & f_{ij}^{\tau+1} \gets f_{ij}^{\tau}+1 \\
& \mathcal{S}_{ij}^{\tau} \leq Top_{R\%}( \mathcal{S}^{\tau}[{1 \dots L_{WH}}]) \\ \\ 
\mathcal{S}_{ij}^{\tau} & f_{ij}^{\tau+1} \leftarrow 0 \\ 
& \mathcal{S}_{ij}^{\tau} >  Top_{R\%}( \mathcal{S}^{\tau}[{1 \dots  L_{WH}}])
\end{cases}
\end{equation*}

where \( \gamma \) is a scaling factor that controls the impact of reuse frequency on the score map \(\mathcal{S}\). \(\tau\) indicates current time step and \(\tau+1\) indicates next time step. \(ij\) indicated index in score map \(\mathcal{S}\) and frequency map \(\mathcal{M}_{freq}\). We have \(\mathcal{M}_{freq}=\{f_{ij},i\in[1\dots L_W],j\in [1 \dots L_H]\}\)

\subsection{Vanilla SLoC w/o ISS and TIP}
\begin{algorithm}[]
\caption{Image Editing with \textbf{SLoC}}
\label{alg:SloC_original}
\begin{algorithmic}[1]
\Require Input image \(\mathbf{I_s}\), Mask for editing region \(\mathbf{M_s}\), Prompt for editing \(\mathbf{P_m}\), Randomly initialized map \(\mathcal{R}\), Bonus for edited region \(\mathbf{S_E}\) and Cache dict \(\mathcal{C}_{l,m}[:]\).
\Ensure The edited result $\mathbf{I}^*$
\State $\mathcal{M}_{freq} \gets \boldsymbol{zero}\mathbf{[\mathcal{F}_i,t,l]}$
\State $\mathbf{Z}_0 \gets \mathbf{cat}(\text{VQ-Encoder}(\mathbf{I_s}),\text{Txt-Encoder}(\mathbf{P_m}))$
\State // \textbf{Image Latent Inversion}
\For{$t = 1, \dots, T$}
    \State $\mathbf{Z}_t \gets \text{RF-inversion}(\mathbf{Z}_{t-1}, t-1, \mathcal{\phi})$
\EndFor
\State \(Z_T^*\gets Z_T\)
\State // \textbf{Image Editing Steps with Caching}
\For{$t = T,T-1, \dots L+1$}
    \For{$\mathcal{F}_i \gets {\mathbf{SA}_l,\mathbf{CA}_l,\mathbf{MLP}_l},l \in [1,\dots\mathcal{L}]$}
    \State \(\mathcal{S}_l \gets  (\mathcal{R} \odot \mathbf{S_E})\oplus \mathcal{M}_{freq} \)
    \State \(\mathcal{I}_{i,l,t} \gets \mathbf{Sel}_{topR\%}(\mathcal{S}_l)\)
    \State \(Z_{l+1}^*\gets \mathbf{scatter}(\mathcal{F}_i(Z_{l}^*,\mathcal{I}_{i,l,t}),\mathcal{C}_{t+1}[l,\mathcal{F}_i])\)
    \State\(\mathbf{Update}(\mathcal{C}_{t+1}[l,\mathcal{F}_i],\mathcal{M}_{freq})\)
    \EndFor
    \State $\mathbf{Z}_{t-1}^* \gets \mathbf{Z}_{t-1}^* \odot \mathbf{M}_s + \mathbf{Z}_{t} \odot (1 - \mathbf{M}_s)$
\EndFor
\State $\mathbf{I}^* \gets \text{VQ-Decoder}(\mathbf{x}_0)$
\State \Return $\mathbf{I}^*$
\end{algorithmic}
\end{algorithm}

\begin{table*}[htbp]
\caption{\textbf{Ablation study for cache}. Comparisons of different cache methods in terms of FG fidelity, and computational efficiency.}
\label{tab:ablation_cache}
\centering
\small
\setlength{\tabcolsep}{4pt}
\scalebox{0.9}{
    \begin{tabular}{lccccccccccc}
    \toprule[1.5pt]
    \multirow{2}{*}{\textbf{Method}} 
     & \multicolumn{4}{c}{\textbf{FG preservation}}
     & \multicolumn{5}{c}{\textbf{Efficiency}} 
     & \multicolumn{2}{c}{\textbf{Speed Up}} 
     \\
    \cmidrule(r){2-5}\cmidrule(r){6-10}\cmidrule(r){11-12}
     & \textbf{FID}$\downarrow$
     & \textbf{PSNR}$\uparrow$
     & \textbf{MSE}$_{10^{-3}}^{\downarrow}$
     & \textbf{SSIM}$_{10^{-1}}^{\uparrow}$
     & \textbf{Steps}
     & \textbf{Inv.(s)}$\downarrow$
     & \textbf{Fwd.(s)}$\downarrow$
     & \textbf{Inference (s)}$\downarrow$
     & \textbf{FLOPs(T)}$\downarrow$ 
     & \textbf{Latency}$\uparrow$
     & \textbf{FLOPs}$\uparrow$ 
     \\
    \midrule
    \textbf{No Cache}  
    & - & - & - & -  
    & 28 & 5.44  & 5.67 & 11.30 & 932.95 & 1 \(\times\) & 1 \(\times\)  \\
    \textbf{50\% Step}  
    & 90.60 & 24.46  & 6.66 & 8.77   
    & 14 & 2.63 & 2.87 & 5.69  & 456.55 & $\fg{2.04\times}$ & $\fg{1.99\times}$  \\
    \textbf{FORA~\cite{fora}}      
    & 39.96 & 31.62  & 1.74 & 9.47  
    & 28 & 2.42 & \textbf{2.45} & 5.05  & 318.09 & $\fg{2.24\times}$ & $\fg{2.93\times}$  \\
    \textbf{ToCa~\cite{toca}}      
    & 84.77  & 26.16   & 5.76 & 8.88   
    & 28 & 3.52 & 3.52 & 7.29  & 332.93 & $\fg{1.55\times}$&$\fg{2.80\times}$ \\ 
    \textbf{DuCa~\cite{duca}}      
    & 84.85 & 26.16  & 5.76 & 8.88   
    & 28 & 3.26 & 3.26 & 6.87  & 313.00 & $\fg{1.67\times}$&$\fg{2.98\times}$ \\
    \bottomrule
    \rowcolor[HTML]{EFEFEF}
    \textbf{SLoC}      
    & 39.50  & \textbf{31.75}  & 1.72 &  \textbf{9.48}   
    & 28 & 2.86 & 2.91 & 5.96  & 384.03 & $\fg{1.90\times}$&$\fg{2.43\times}$ \\
    
    \rowcolor[HTML]{EFEFEF}
    \textbf{SLoC\tiny{+TIP+ISS}}      
    & \textbf{39.21}  & \textbf{31.75}  & \textbf{1.71} & \textbf{9.48}   
    & 28 & \textbf{1.92} & 2.49 & \textbf{4.60}  & \textbf{264.50} & $\fg{\textbf{2.46}\times}$ & $\fg{\textbf{3.53}\times}$  \\
    
    \bottomrule[1.5pt]
    \end{tabular}
}

\end{table*}
\section{Discussion on TIP}
We adopt Token Index Preprocessing.
This design offers an additional advantage by reducing the number of function calls within the cache module. Specifically, with a preprocessing overhead of no more than \textbf{150ms}, we achieve a reduction of over \textbf{1000ms} in cache-induced inference latency. 
Since the token selection in our algorithm is independent of the internal properties of individual tokens or their mutual interactions, this decoupling is logically equivalent in the temporal sequence. Consequently, it enables further lossless acceleration on top of SLoC.

\subsection{Proof of TIP Equivalence with Original Operations}
\label{proof}
We maintain a cache of intermediate features for a set of tokens in SLoC. At each iteration step, each token is assigned a score based on (i) a random or seed-based component and (ii) a function of its selection frequency (the \(\mathcal{M}_{freq}\)). The top $R\%$ of tokens are selected for updating the cache. It follows algorithm~\ref{alg:SloC_original}.

We prove that this preprocessing-based approach is mathematically equivalent to performing scoring, sorting, and token selection \emph{online} at each iteration.

\subsection{Notation and Problem Setup}
\begin{itemize}
    \item $N$: Total number of tokens, indexed as $\{1,2,\dots,N\}$.
    \item $T$: Total number of diffusion iterative steps.
    \item $s_i^{(t)}$: Score of token $i$ at step $t$, given by
    \[
        s_i^{(t)} = f(r_i^{(t)}) + \mathcal{M}_{freq,i}^{(t)}
    \]
    where:
    \begin{itemize}
        \item $r_i^{(t)}$: Random (or seed-based) component.
        \item $\mathcal{M}_{freq,i}^{(t)}$: Frequency of times token $i$ has been selected before step $t$.
        \item $f(\cdot)$: A deterministic function adjusting scores, region score bonus adopted in SLoC here.
    \end{itemize}
    \item After computing $\{s_i^{(t)}\}_{i=1}^N$, the top $R\%$ tokens are selected for cache updates.
    \item For simplicity in our proof, we have omitted the layer index and module type (Cross-Attention, Self-Attention, MLP).
\end{itemize}

\subsection{Original (Online) Algorithm Description}
The online method iterates as follows~\ref{alg:SloC_original}:
\begin{enumerate}
    \item For $t = 1$ to $T$:
    \begin{enumerate}
        \item Compute $s_i^{(t)}$ for each token $i$.
        \item Sort tokens by $s_i^{(t)}$ and select the top $R\%$.
        \item Update the cache for these selected tokens.
        \item Increment $\mathcal{M}_{freq,i}^{(t+1)}$ for each selected token $i$.
    \end{enumerate}
\end{enumerate}
Here, $r_i^{(t)}$ is reproducible when using a fixed seed.

\subsection{Proposed Optimization (TIP)}
The optimized approach precomputes the cache update and selection process~\ref{alg:edit}:
\begin{enumerate}
    \item Generate and iterate all $r_i^{(t)}$ for $i = 1,\dots,N$ and $t = 1,\dots,T$.
    \item Simulate the selection process offline:
    \begin{enumerate}
        \item Initialize $\mathcal{M}_{freq,i}^{(1)}=0$ for all $i$.
        \item For each $t = 1$ to $T$:
        \begin{itemize}
            \item Compute $ s_i^{(t)} = f(r_i^{(t)}) + \mathcal{M}_{freq,i}^{(t)}$.
            \item Sort and select the top $R\%$, recording indices as $\mathcal{I}^{(t)}_{\text{top}}$.
            \item Update $\mathcal{M}_{freq,i}^{(t+1)}$ for selected tokens.
        \end{itemize}
    \end{enumerate}
    \item Store $\{\mathcal{I}^{(t)}_{\text{top}}\}_{t=1}^T$ for later use.
\end{enumerate}
At inference, we read precomputed $\mathcal{I}^{(t)}_{\text{top}}$ instead of recomputing scores.

\subsection{Proof of Equivalence}
We prove that both methods select identical tokens at each step.

\paragraph{Step 1 Equivalence.} 
At $t=1$, we have $\mathcal{M}_{freq,i}^{(1)}=0$, so
\[
   s_i^{(1)} = f(r_i^{(1)}) + 0.
\]
Since $r_i^{(1)}$ is identical in both methods (fixed seed), sorting $s_i^{(1)}$ gives the same top $R\%$ tokens, ensuring identical updates and increments for $\mathcal{M}_{freq,i}^{(2)}$.

\paragraph{Inductive Hypothesis.} 
Assume for steps $k < t$ that
\[
   \mathcal{I}^{(k)}_{\text{top}}(\text{offline}) = \mathcal{I}^{(k)}_{\text{top}}(\text{online}).
\]
Thus, $\mathcal{M}_{freq,i}^{(t)}$ is identical in both methods.

\paragraph{Step $t$ Equivalence.} 
At step $t$,
\[
   s_i^{(t)} = f(r_i^{(t)}) + \mathcal{M}_{freq,i}^{(t)}
\]
Since $r_i^{(t)}$ and $\mathcal{M}_{freq,i}^{(t)}$ are identical (by induction), we get
\[
   s_i^{(t)}(\text{offline}) = s_i^{(t)}(\text{online}),
\]
ensuring that sorting and selecting the top $R\%$ gives identical indices sets:
\[
   \mathcal{I}^{(t)}_{\text{top}}(\text{offline}) = \mathcal{I}^{(t)}_{\text{top}}(\text{online}).
\]

\paragraph{Conclusion by Induction.} 
By induction, token selection and cache updates remain identical for all $t=1,\dots,T$. Thus, preprocessing achieves the same outcome as the online approach.




\section{Metrics}
Our experiments employ a selection of the most widely used image quality, instruction adherence, and efficiency metrics.

\textbf{Frechet Inception Distance (FID)} and \textbf{Learned Perceptual Image Patch Similarity (LPIPS)} are feature-based similarity metrics computed using pretrained neural networks. Lower values indicate higher similarity. We use \texttt{InceptionV3} for FID and \texttt{AlexNet} for LPIPS measurements.
\textbf{Peak Signal-to-Noise Ratio (PSNR)} and \textbf{Mean Squared Error (MSE)} are pixel-space similarity metrics. A higher PSNR and a lower MSE indicate greater image similarity.
\textbf{CLIPScore} measures how well an image generation or editing result follows a given prompt using a pretrained CLIP model. A higher score indicates stronger adherence to the prompt. In our experiments, we use the \texttt{clip-vit-base-patch16} model.
\textbf{FLOPs} quantify the computational cost associated with model inference. A higher value indicates greater computational overhead.

\section{More Experiments}
\label{appendix_exp}
\begin{table}[ht]
    \centering
    \caption{{\textbf{Performance comparison}. An ablation study is conducted on imbalanced inversion and denoising for background preservation, foreground fidelity and inference time.}}
    \label{tab:ablation_imbalanced}
    \scalebox{0.6}{
        \begin{tabular}{cc|c|ccc|c}
                    \toprule
                    \multirow{3}{*}{\textbf{Inversion}} & 
                    \multirow{3}{*}{\textbf{Denoising}}  & 
                    \multicolumn{1}{c|}{\textbf{BG Preservation}} & 
                    \multicolumn{3}{c|}{\textbf{FG Fidelity}} & 
                    \multirow{2}{*}{\textbf{Inference} $\uparrow$} \\
                    \cmidrule(lr){3-3} \cmidrule(lr){4-6}
                    & & \textbf{LPIPS} $_{\times 10^{-2}}^\downarrow$ & \textbf{LPIPS} $_{\times10^{-3}}^\downarrow$ & \textbf{PSNR $\uparrow$} & \textbf{FID$\downarrow$} &  Time (s) \\
                    \bottomrule
                    \midrule
                    \textbf{Full Step} & \textbf{Full Step} & 1.98 & - & - & - & 13.27 \\
                    \midrule
                    2-step skip & \multirow{3}{*}{\textbf{Full Step}} & 31.38 & 1.98 & 31.93 & 3.35 & 10.16 \\
                    3-step skip&  & 31.38 & 1.98 & 25.79 & 3.23 & 9.31  \\
                    4-step skip&  & 31.38 & 1.98 & 26.50 & 3.31 & 8.76  \\
                    \midrule
                    \multirow{3}{*}{\textbf{Full Step}} 
                    & 2-step skip & 1.97 & 50.40 & 31.93 & 28.51  & 11.79  \\
                    & 3-step skip & 1.97 & 121.44 & 25.79 & 64.10  & 9.28  \\
                    & 4-step skip & 1.96 & 102.67 & 26.50 & 56.93  & 8.54  \\
                    \bottomrule
        \end{tabular}
    }
    
\end{table}

We compared the edited results in terms of image similarity and efficiency. The Table.~\ref{tab:ablation_cache} demonstrate that our approach, when incorporating all optimizations (TIP + ISS), achieves the best performance in both fidelity and efficiency.
SLoC achieves a \textbf{2.46}$\times$ significant improvement in inference latency and a \textbf{3.53}$\times$ acceleration in computational efficiency compared to the original unaccelerated version. Additionally, our method demonstrates either improved fidelity or remains at a state-of-the-art level across various editing tasks.

\section{Related work}
We thanks the related work about editing both image and video, such as VideoGrain~\cite{yang2025videograin}, FastVAR~\cite{guo2025fastvar}, IntLora~\cite{guo2024intlora}, FollowFamilty~\cite{ma2024followyouremoji,ma2025followyourclick,ma2022visual,ma2024followpose,ma2025followcreation,ma2025followyourmotion,yan2025eedit,zhang2025magiccolor,zhu2024instantswap,wang2024cove,feng2025dit4edit,chen2024follow}, Tpsence~\cite{zheng2024tpsence}, Pointnorm~\cite{zheng2023pointnorm}, MotionDiff~\cite{ma2025motiondiff}, FastScene~\cite{ma2024fastscene}, DreamRelation~\cite{wei2025dreamrelation}, Dreamvideo~\cite{wei2024dreamvideo}, and Lazymar~\cite{yan2025lazymar} and so on ~\cite{chen2025transanimate,song2022cliptexture,song2023clipvg,song2022clipfont}.
\begin{figure}
    \centering
    \includegraphics[width=\linewidth]{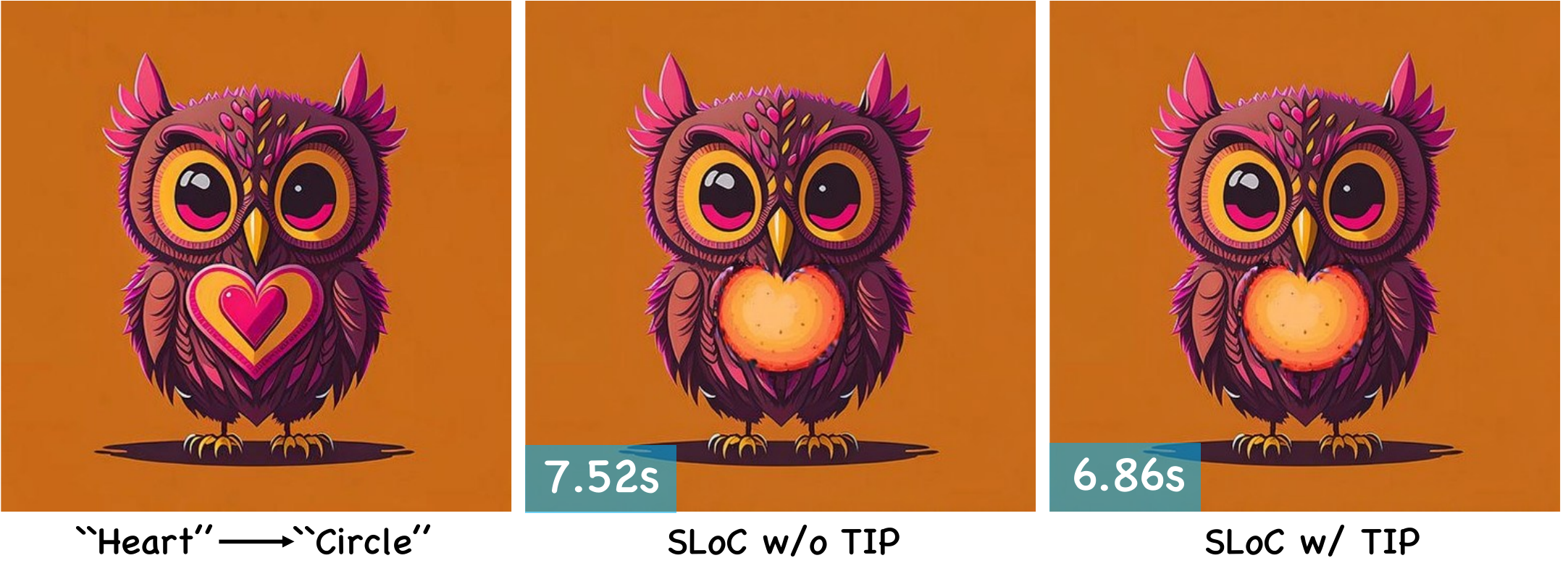}
    \caption{A qualitative example. Token index preprocessing shows loss-less acceleration for editing quality.}
    \label{fig:ablation_tip}
\end{figure}

\subsection{AI safety discussion}
As discussed in previous works such as watermark techniques~\cite{song2025idprotector,song2024anti,hui2025autoregressive,ci2024wmadapter,ci2024ringid,liu2024image,yang2024can}, AI-generated content should be more responsible and safe by adding watermark sign. We will release relative code to make sure our work be safe and not serve for harmful circumstance.

\subsection{More relative controllable methods}
In more recent works, controllable generation of diffusion models remains a popular topic, such as
~\cite{zhang2025easycontrol,zhang2024ssr,song2025layertracer,song2025makeanything,huang2025photodoodle,song2025omniconsistency,guo2025any2anytryon,song2024diffsim,song2024processpainter,zhang2025stable,zhang2024stable,wan2024grid,wang2025diffdecompose,gong2025relationadapter,jiang2025personalized,lu2025easytext,shi2024fonts,shi2025wordcon}
These works inspire us to apply locally sensitive caching in a wider range of controllable generation and editing tasks to achieve faster generation acceleration.

\section{Gallery}
We present additional editing results, including prompt-guided, drag-guided, and reference-guided editing. Furthermore, we adapted the community-developed Redux model for img2img tasks, enabling the generation of impressive image variations.

\begin{figure*}
    \centering
    \includegraphics[width=\linewidth]{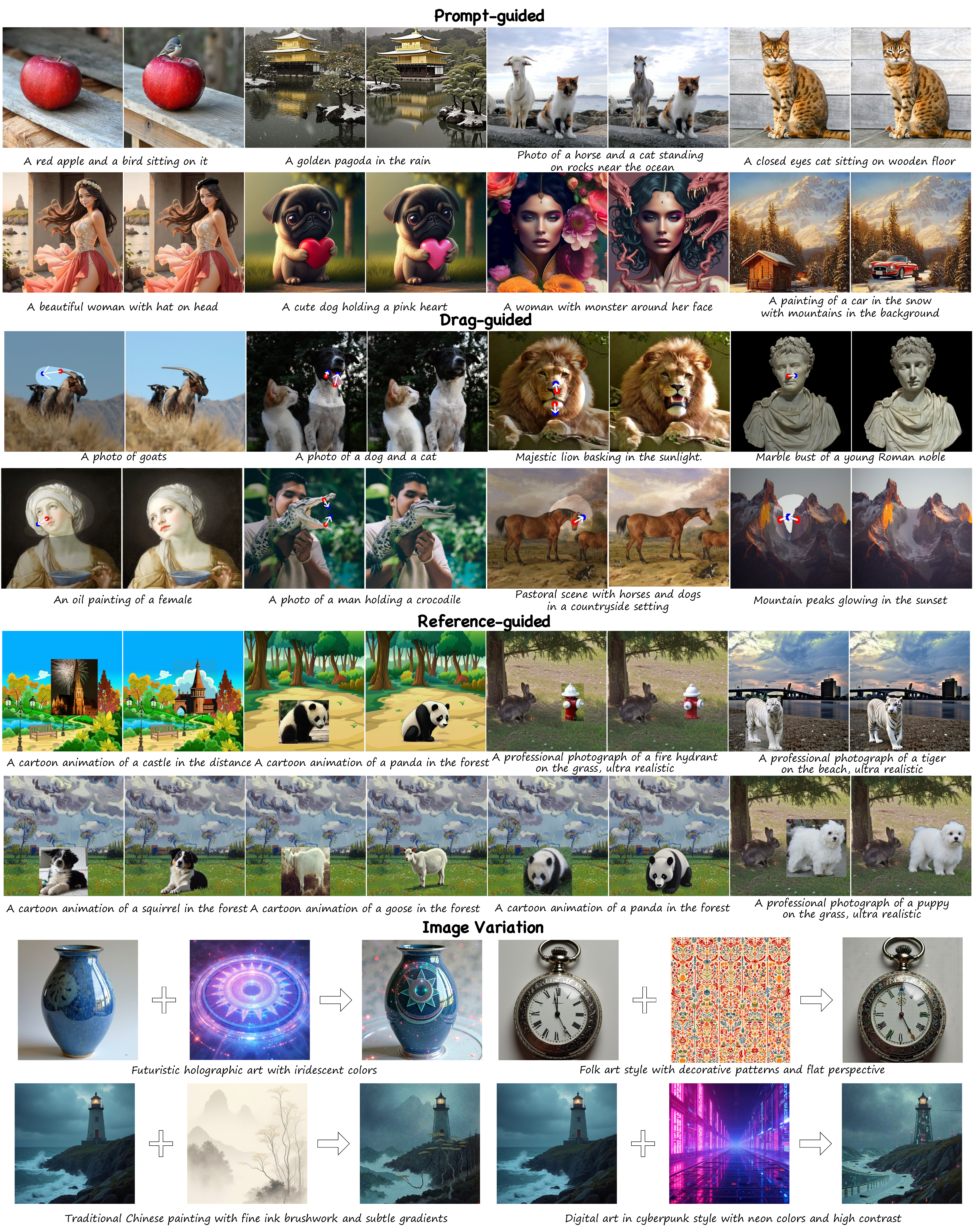}
\end{figure*}
\begin{figure*}
    \centering
    \includegraphics[width=\linewidth]{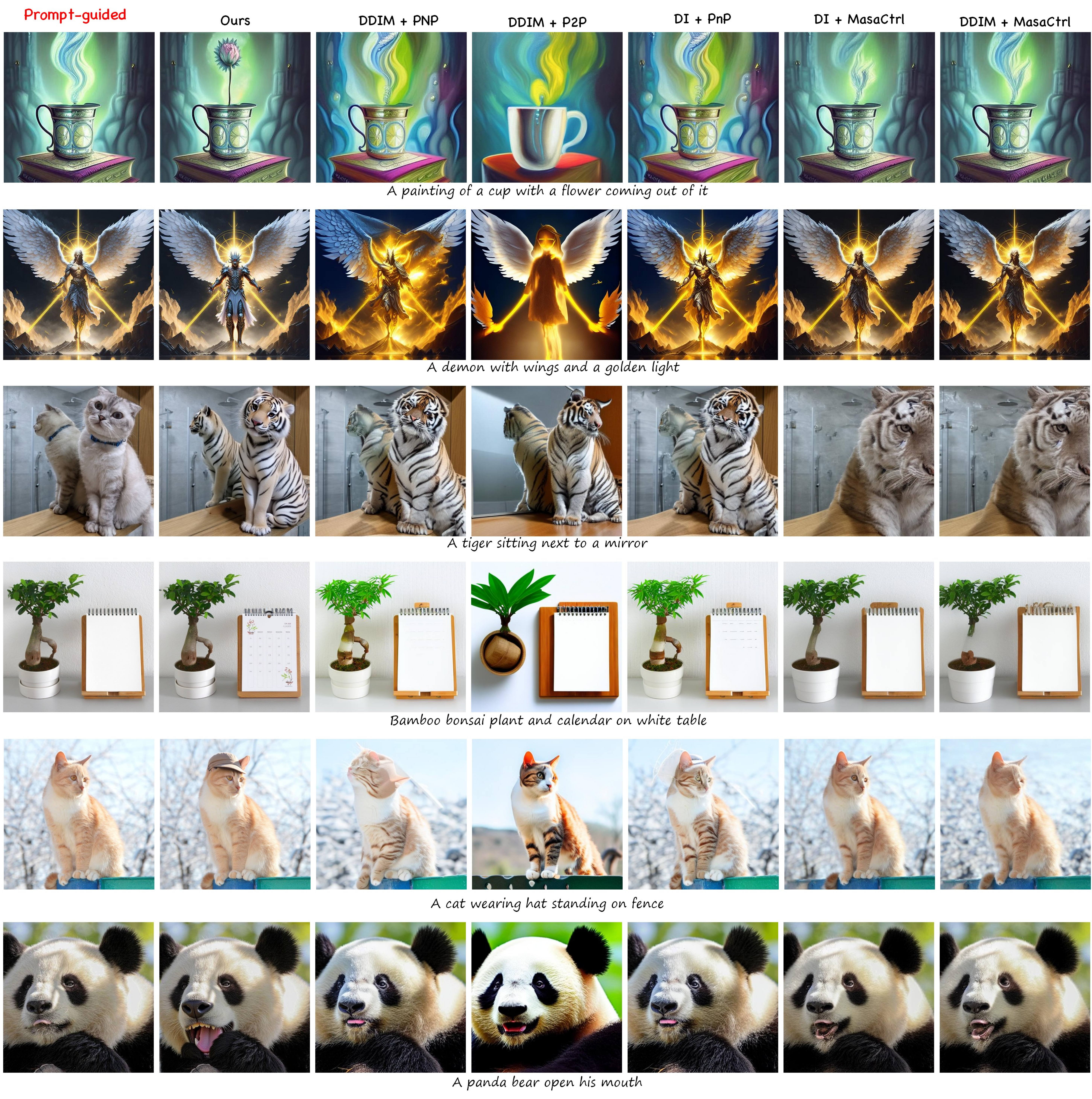}
\end{figure*}
\begin{figure*}
    \centering
    \includegraphics[width=\linewidth]{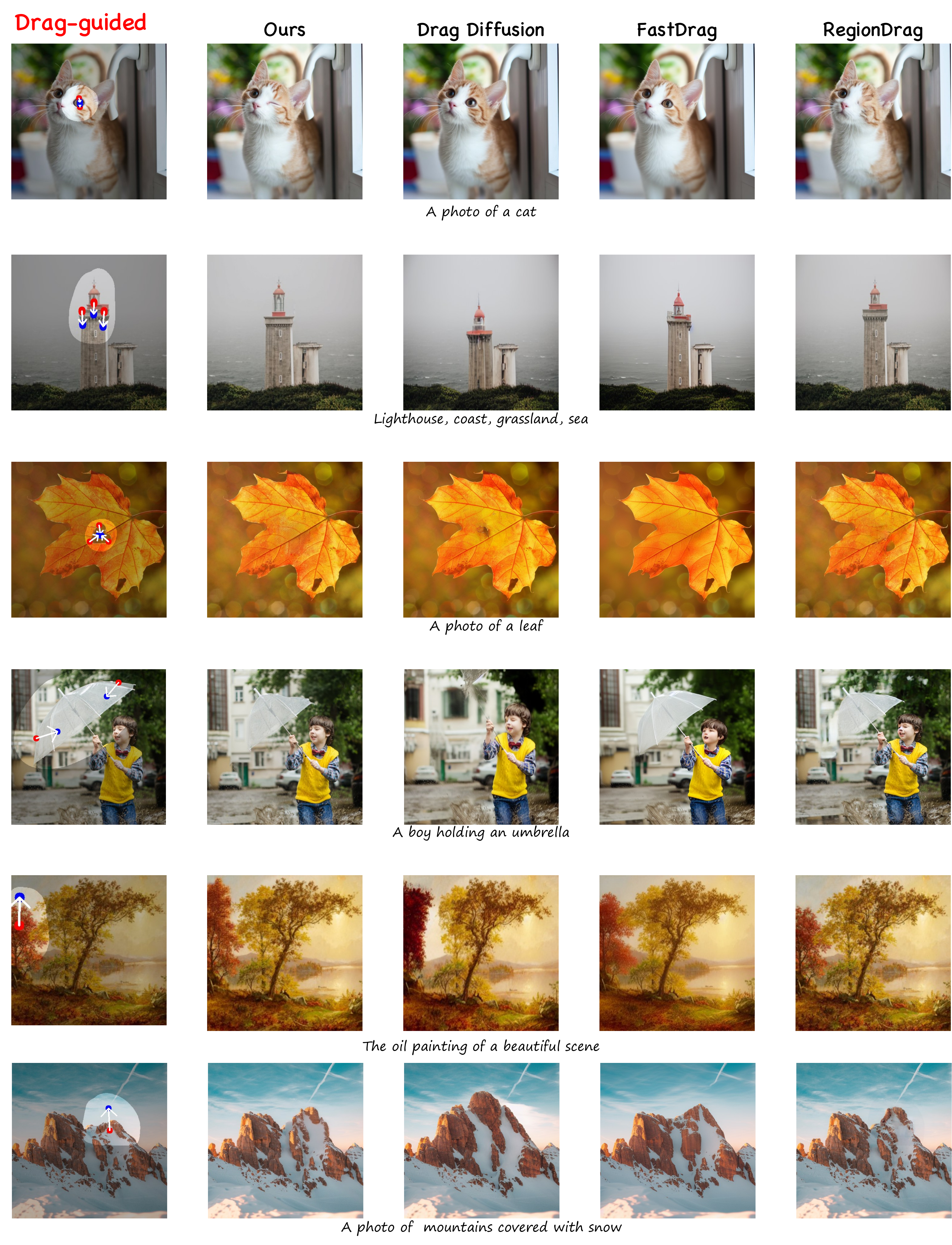}
\end{figure*}

\begin{figure*}
    \centering
    \includegraphics[width=\linewidth]{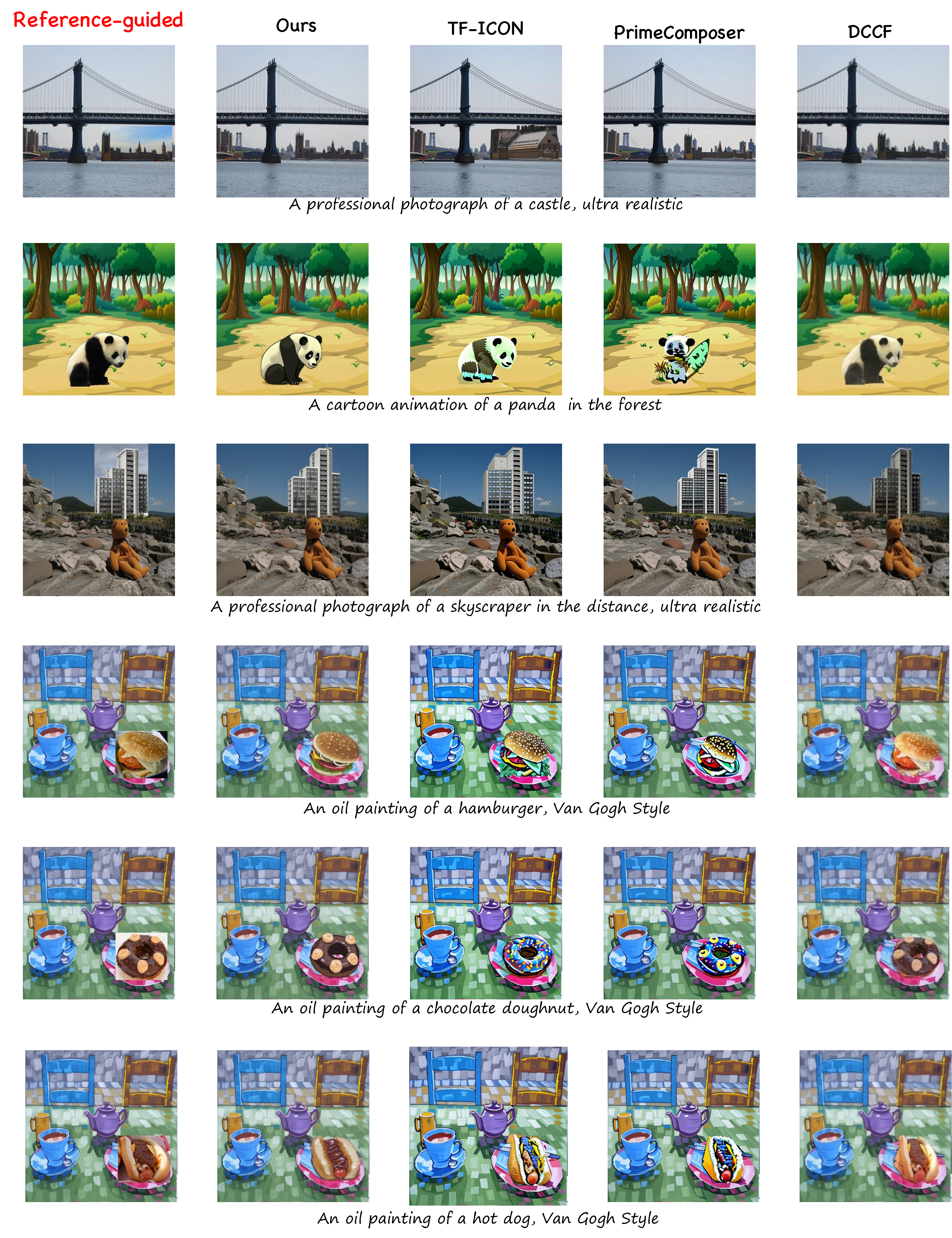}
\end{figure*}